\documentclass[pdflatex,sn-mathphys-num]{sn-jnl}

\usepackage{graphicx}%
\usepackage{multirow}%
\usepackage{amsmath,amssymb,amsfonts}%
\usepackage{amsthm}%
\usepackage{mathrsfs}%
\usepackage[title]{appendix}%
\usepackage{xcolor}%
\usepackage{textcomp}%
\usepackage{manyfoot}%
\usepackage{booktabs}%
\usepackage{algorithm}%
\usepackage{algorithmicx}%
\usepackage{algpseudocode}%
\usepackage{listings}%
\usepackage{geometry}
\newcommand{\indep}{\perp \!\!\! \perp}

\usepackage{longtable}
\usepackage{caption}
\usepackage{multirow}
\usepackage{makecell}
\usepackage{tablefootnote}
\usepackage{url}
\usepackage{soul} 
\begin{document}

\title[Article Title]{Who cuts emissions, who turns up the heat? causal machine learning estimates of energy efficiency interventions}

\author*[1,2]{\fnm{Bernardino} \sur{D'Amico}}\email{b.damico@napier.ac.uk}
\author[3]{\fnm{Francesco} \sur{Pomponi}}\email{francesco.pomponi@york.ac.uk}
\author[4]{\fnm{Jay} \sur{H. Arehart}}\email{jay.arehart@colorado.edu}
\author[1,2]{\fnm{Lina} \sur{Khaddour}}\email{l.khaddour@napier.ac.uk}

\affil*[1]{\orgdiv{Digital Built Environment Group (DiBEG)}, \orgname{Edinburgh Napier University}, \orgaddress{\city{Edinburgh}, \postcode{EH10 5DT}, \country{United Kingdom}}}

\affil[2]{\orgdiv{School of Computing, Engineering and the Built Environment (SCEBE)}, \orgname{Edinburgh Napier University}, \orgaddress{\city{Edinburgh}, \postcode{EH10 5DT}, \country{United Kingdom}}}

\affil[3]{\orgdiv{York School of Architecture}, \orgname{University of York}, \orgaddress{\city{York}, \postcode{YO10 5DD}, \country{United Kingdom}}}

\affil[4]{\orgdiv{Department of Civil, Environmental, and Architectural Engineering}, \orgname{University of Colorado Boulder}, \orgaddress{\city{Boulder}, \postcode{80309-0428}, \state{Colorado}, \country{United States}}}

\abstract{Reducing domestic energy demand is central to climate mitigation and fuel poverty strategies, yet the impact of energy efficiency interventions is highly heterogeneous. Using a causal machine learning model trained on nationally representative data of the English housing stock, we estimate average and conditional treatment effects of wall insulation on gas consumption, focusing on distributional effects across energy burden subgroups. While interventions reduce gas demand on average ($\approx$–19\%), low energy burden groups achieve substantial savings, whereas those experiencing high energy burdens see little to no reduction. This pattern reflects a behaviourally-driven mechanism: households constrained by high costs-to-income ratios ($>$10\%) reallocate savings toward improved thermal comfort rather than lowering consumption. Far from wasteful, such responses represent rational adjustments in contexts of prior deprivation, with potential co-benefits for health and well-being. These findings call for a broader evaluation framework that accounts for both climate impacts and the equity implications of domestic energy policy.}


\maketitle

\section{Introduction}\label{sec_main}

Domestic energy use accounts for a substantial portion of global greenhouse gas (GHG) emissions, making its reduction a cornerstone of climate mitigation policies worldwide  \cite{barrett2022energy, xia2024comparative}. In pursuit of decarbonisation, governments and international authorities have mandated laws, standards, subsidies and several other programmes and incentives to improve on energy efficiency of the domestic building stock \cite{broin2015energy, economidou2020review, mcandrew2021household}. As such, energy efficiency remains a critical pillar in these efforts for the foreseeable future, both as a means of reducing GHG emissions, and as a cost-effective strategy to accelerate progress toward national and international net-zero targets set for mid-century milestones \cite{iea2022netzero, ipcc2023synthesis, uknetzero2050}. 

Improving energy efficiency in residential buildings is widely perceived as a particularly attractive climate policy goal because it promises to deliver two key objectives at once: lower GHG emissions and improve household wellbeing \cite{ballesteros2022effects, mei2024relationships}. This latter promise rests on the expectation that efficiency upgrades make it easier and more affordable to maintain adequate indoor temperatures, thereby reducing exposure to cold, damp, and fuel-related financial strain ---all of which have been closely linked to respiratory illness \cite{moorcroft2025damp}, cardiovascular conditions \cite{singh2022estimating}, anxiety and depression \cite{harris2010health} and more generally identified as determining factor of excess winter deaths \cite{ons2021excess, lee2022fuel}. This situation is commonly referred to as \emph{fuel poverty}, defined as a household’s inability to afford adequate energy services to maintain a healthy and comfortable living environment. 

As such, policy co-benefits expected from energy efficiency interventions \cite{haines2009public} are further reinforced by the broader health advantages of reduced fossil fuel use, including improved air quality and reduced pressure on public healthcare.

Existing research has shown however that domestic energy efficiency interventions often lead to highly heterogeneous cross-population reductions in energy demand \cite{mccoy2021quantifying, penasco2023assessing}. In many cases, the actual energy savings fall short of the projections made by engineering evaluations based on building physics models ---a phenomenon commonly referred to as the \emph{rebound} effect \cite{aydin2017energy}. 
While the rebound effect was initially dismissed as technical anomaly, it has been increasingly attributed to behavioural factors \cite{cali2016energy}.

Notably, socioeconomic factors (particularly income) appear to play a central role in shaping post-retrofit energy behaviour \cite{mccoy2021quantifying}, with higher-income households tending to achieve stronger and more sustained reductions in energy consumption following efficiency upgrades. In contrast, lower-income households often exhibit considerably weaker, and in some cases negligible, reductions in energy use \cite{penasco2023assessing} post intervention.

While behavioural factors undoubtedly influence how households respond to energy efficiency interventions, a robust causal explanation must also consider the extent to which these behaviours are shaped by the intervention itself. A central mechanism in this regard is the \emph{prebound} effect \cite{sunikkablank2012prebound}, where pre-intervention energy use is lower than engineering model estimates due to suppressed heating behaviours, as a result of affordability constraints. This effect is particularly pronounced among fuel-poor households, who may limit heating to unsafe or unhealthy levels in order to manage high energy costs \cite{anderson2012coping, san2018relationship}. 
For instance, recent ONS data report that a substantial proportion of UK households refrain from adequately heating their homes even in cold conditions; as of late 2022, more than 6 out of 10 (63\%) adults indicated they were consuming less gas and electricity due to rising living costs, with over 9 out of 10 (96\%) of these individuals specifically reducing their heating frequency \cite{ons_winter_2022}.

In such cases, it is plausible that energy efficiency upgrades do not eliminate `excess' energy use but instead enable a shift toward more adequate and healthy indoor conditions. While the prebound effect has been documented in observational studies, robust quantification of its causal impact across heterogeneous households remains limited. Our study contributes by estimating these causal effects using a causal machine learning framework.

Building on previous work \cite{mccoy2021quantifying}, we adopt the hypothesis that when a dwelling undergoes an energy efficiency upgrade which, in addition to reducing required energy use, also lowers associated energy costs, this may induce behavioural changes to household heating patterns. Specifically, the hypothesis implies that exactly the same intervention alleviates energy burden (i.e., the proportion of income spent on energy bills) much more significantly, in relative terms, for lower-income households. As a result, these households are more likely to reallocate the meaningful financial relief gained from efficiency improvements toward increased thermal comfort rather than maximising energy savings, given that their heating behaviour was previously constrained to suboptimal comfort levels \cite{anderson2012coping} due to high energy burden pressures. As such, this behaviourally-driven energy rebound is consistent with rational household responses under prior deprivation, highlighting the need to account for heterogeneity in evaluating intervention outcomes.

Previous attempts to infer the causal impact of energy efficiency interventions on household energy use rely on observational data sourced from the National Energy Efficiency Data (NEED) framework and employ econometric techniques such as difference-in-differences and statistical matching \cite{mccoy2021quantifying, penasco2023assessing}. The NEED dataset is particularly valuable for its large-scale, longitudinal records of meter-level energy consumption. However, it lacks detailed socioeconomic data (most notably, income) at household-unit level. Furthermore, it covers records of energy efficiency interventions carried out through government schemes, to which households (or their landlord, social or private) have voluntarily adhered to. As a result, treatment assignment in the NEED sample dataset is subject to potential self-selection bias \cite{heckman1990selection} i.e. households that choose to participate in retrofit programmes may differ systematically from non-participants in ways that also affect their energy consumption patterns, thus limiting the generalisability and internal validity of causal estimates based solely on this dataset.

Building on established findings regarding prebound and rebound effects, our study introduces two key methodological and data-driven innovations. First, we develop a causal graphical model \cite{pearl2009causality} ---a technique within the broader field of so-called causal machine learning (ML), whose aim is to estimate causal effects rather than mere predictions from observed data. By using a (causal) graphical model approach we are able to explicitly formalises the assumptions underlying our causal effect identification strategy, hence obtaining estimations of both average treatment effects (ATE) and conditional average treatment effects (CATE). This approach offers greater transparency than conventional matching or difference-in-differences techniques, especially when addressing complex confounding and heterogeneity.
Second, we source data from the English Housing Survey Fuel Poverty (EHS-FP) dataset series, which provides detailed unit-level information on household income, housing characteristics as well as modelled energy costs. While the reliance on modelled (rather than metered) energy costs introduces a degree of uncertainty ---a limitation we directly address in section \ref{sec:datasourcing}--- the EHS-FP includes nationally representative sampling weights for each unit. This allows us to generalise our findings to the entire stock of over 22 million dwellings in England. Crucially, the availability of directly observed household income at the unit level enables precise calculation of fuel burden, in contrast to reliance on indirect proxies. The only two existing works attempting to identify effect heterogeneity of energy efficiency interventions across household types  \cite{mccoy2021quantifying, penasco2023assessing} rely on area-level indicators such as the Index of Multiple Deprivation (IMD), which fails to capture important within-area income heterogeneity \cite{clelland2019deprivation, deas2003measuring}.

\section{Methods}\label{sec_methods}
Our methodology is grounded in the concept of \emph{probabilistic causation} \cite{suppes1973probabilistic}, thus interpreting an external intervention as affecting the change in probability for an outcome to occur, rather than as a deterministic mechanism. Specifically, we adopt a causal graphical model approach \cite{pearl2009causality}, in which the system of interest is represented by a directed acyclic graph (DAG) that encodes assumed causal relationships among variables. Each node in the DAG corresponds to a random variable, and arrowed edges represent direct causal influences between variable pairs (Figure \ref{fig:main_graphs}).

This formalism relies on two key assumptions: the \emph{causal Markov} condition (Section \ref{Sec:modularity}) and the \emph{faithfulness} condition. The latter ensures that statistical independencies implied by the DAG structure are reflected in the observed data. Together, these assumptions enable the use of the graph to identify appropriate adjustment sets and to infer causal effects from observational distributions.

Causal graphical models offer a systematic and transparent framework for causal inference from observational data, which is particularly valuable in settings where randomised controlled trials are unethical, infeasible (as in this case), or prohibitively expensive. Although the formalisation of graph-based causal inference emerged from the field of artificial intelligence ---most notably through the foundational work of Pearl \cite{pearl2009causality}--- the framework has since been widely adopted across diverse disciplines concerned with causal effect estimation. In epidemiology and public health research for instance, DAGs are established tools to represent confounding, identify adjustment variables, and support causal claims \cite{greenland1999causal, griffith2020collider, tennant2021use}. Similarly, in econometrics, graph-based causal reasoning has become an established tool for policy evaluation and treatment effect estimation \cite{spirtes2005graphical, heckman2024econometric, hunermund2025causal}.

Building on these foundations, recent developments of causal ML have extended the principles of graphical modelling to high-dimensional and data-rich environments. For example, in the Earth system sciences, causal discovery methods have been used to identify drivers of complex climate phenomena from observational time series \cite{runge2019inferring}. In healthcare and medicine, causal ML has improved diagnostic accuracy and personalised treatment strategies beyond correlational approaches \cite{richens2020improving, sanchez2022causal, feuerriegel2024causal} such as those based on deep (yet opaque \cite{pearl2019limitations}) neural networks. More recently, foundation models to predict missing values in tabular data have shown predictive performance beyond current state-of-the-art when based on graphical (structural) causal models \cite{hollmann2025accurate}.

These methodological foundations provide the framework we relied upon to analyse the causal relationship between insulation interventions, household energy use, and behavioural adjustments. Before specifying the causal graph, we first describe the data on which the model is trained and the procedures used to prepare it.

\subsection{Data sourcing and preparation}
\label{sec:datasourcing}

Data used to train the causal graphical model were obtained from the English Housing Survey - Fuel Poverty (EHS-FP) dataset collection \cite{EHS_FP_2018}, a yearly, nationally representative survey released by the UK government that collects information on household and dwelling characteristics necessary for assessing fuel poverty in England.
To ensure methodological consistency, we selected EHS-FP datasets from the pre-pandemic period (2015–2018). During this period, data collection followed standard procedures involving in-person interviews and property inspections, ensuring comparability across survey years.
Each yearly EHS-FP dataset contains between 11,853 and 12,024 household records, with the exact number varying by reference year. Approximately 16\% of these records were excluded for not listing gas as the main fuel type. Each remaining household record includes a sampling weight variable \cite{UKDA9243}, which indicates the household's representativeness within the national domestic building stock population of approximately 22.6 million dwellings. These weights were used to perform stratified resampling, generating a weighted dataset of 60,000 household units for each of the four yearly datasets. Testing larger resampled datasets showed no appreciable difference in the estimated results. The final dataset used for training the model parameters therefore consists of 240,000 weighted units in total and is representative of the English domestic building stock across the years 2015–2018.
Compared to other datasets such as NEED (which tracks households participating in energy efficiency programmes), the EHS-FP provides a randomly selected sample of the overall (English) housing stock. NEED is longitudinal but primarily includes programme participants, potentially introducing self-selection bias, whereas the EHS-FP provides a representative baseline for country-wide causal inference.

A detailed description of all variables, including definitions and measurement units, is provided in section \ref{subsec_variables_descr} of the supplementary material.

\subsection{Causal graph specification}
\begin{figure}[ht]
    \centering
    \includegraphics[width=14.8cm]{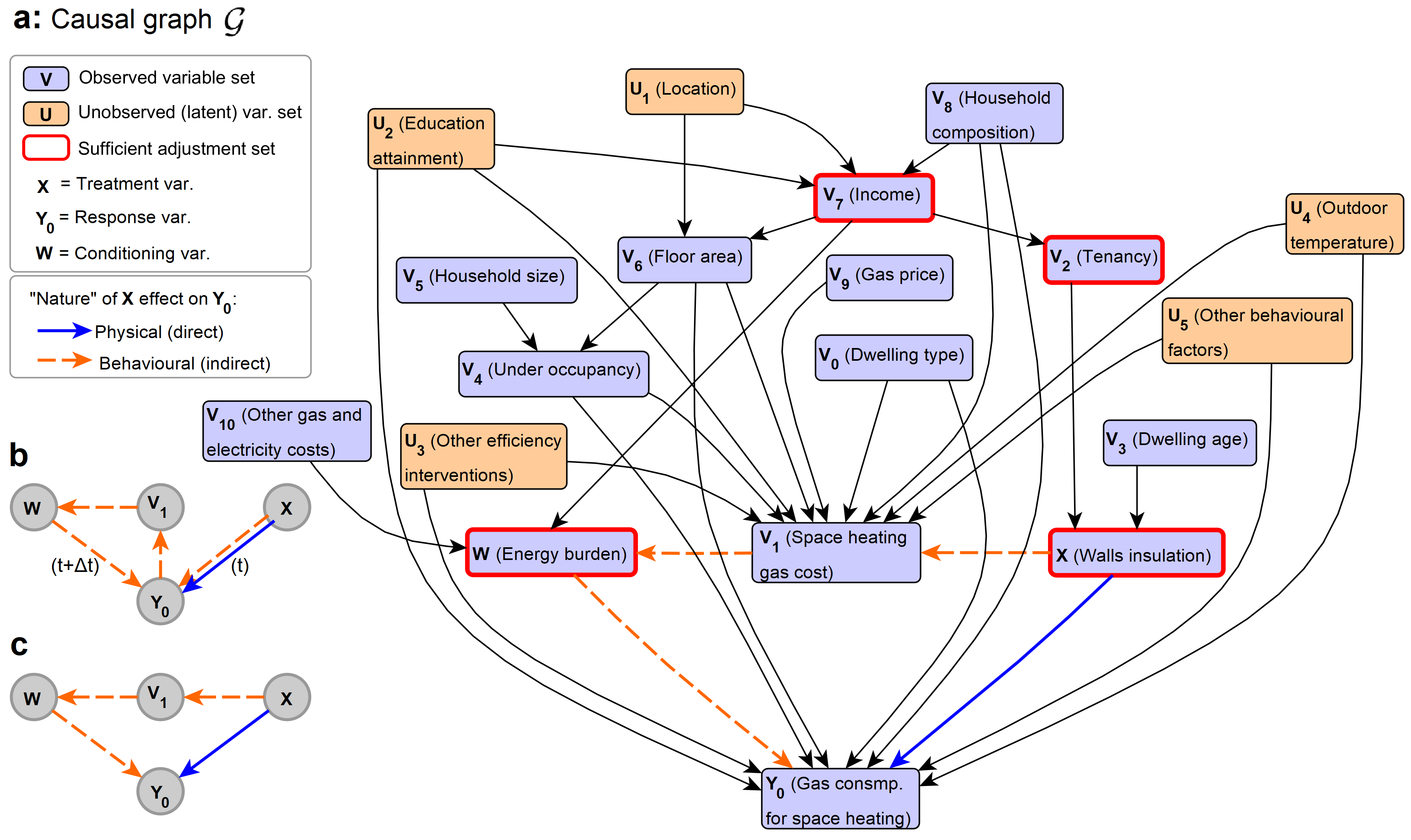}
    \caption{\small Causal graphs specification representing the structure of assumed causal relationships. \textbf{a}: full directed acyclic graph (DAG) $\mathcal{G}$, incorporating the intervention variable, $X$, the outcome, $Y_0$, and additional covariates representing potential confounders and mediators such as energy burden, $W$. \textbf{b}: the core causal mechanism is first modelled as a cyclic temporal graph, thus capturing the dynamic feedback loop initiated by the intervention $X$ and mediated through gas consumption $Y_0$ at time $t$, then gas cost ($V_1$), and household energy burden ($W$), eventually influencing future gas consumption at time $t+\Delta t$. \textbf{c}: modified acyclic version of the core causal graph, derived by unfolding the feedback loop into two temporally undifferentiated causal pathways, i.e. a direct physical effect ($X \rightarrow Y_0$) and an indirect behavioural effect ($X \rightarrow V_1 \rightarrow W \rightarrow Y_0$), as formalised in Eq. (\ref{two_chains}). This transformation from cyclic to acyclic ensures compatibility with the requirement of DAG-based causal inference modelling.}
    \label{fig:main_graphs}
\end{figure}

The DAG constructed for our model is shown in Figure \ref{fig:main_graphs}. Here, the $X$ symbol designates the intervention variable \emph{walls insulation}, whereas $Y_0$ and $W$ respectively indicate the outcome variable \emph{gas consumption for space heating} and the covariate \emph{household energy burden}. This latter is taken as the ratio between gas costs for space heating $V_1$ (plus all other gas and electricity costs, $V_{10}$), and household income $V_7$, i.e. $w=(v_1+v_{10})/v_7$.

To formally describe the core causal mechanism from a temporal perspective, we begin with the following causal sequence: 

\begin{equation}
    \label{behavioural_chain}
    X \rightarrow Y_0^{t} \rightarrow V_1 \rightarrow W \rightarrow Y_0^{t+\Delta t}
\end{equation}
illustrating how an external intervention entailing an energy efficiency measure to the dwelling $X=x$ ---i.e. installation of external walls insulation in the specific case--- initially leads to a reduction in energy (gas) consumption $Y_0$ at time $t$. This decrease in consumption subsequently lowers gas cost expenditures ($V_1$), thereby reducing the household's energy burden ($W$). The change in energy burden, in turn, influences occupants behaviour, potentially affecting subsequent gas consumption ($Y_0$) at a later time $t+\Delta t$. 

Following the initial reduction in gas consumption at $t$, there may be a compensatory behavioural response leading to an increase in consumption at $t+\Delta t$. Consequently, the causal effect triggered by $X$ enters a feedback loop involving the variables $Y_0$, $V_1$, $W$ ultimately returning to $Y_0$. As a result, both gas consumption and associated costs oscillate around (and eventually settle on) an equilibrium state post-intervention, which is constrained by the governing price relationship correlating cost and consumption of gas for a household:

\begin{itemize}
    \item[] \emph{Gas cost} ($V_1$) [£] $\approx$\emph{Gas consumption} ($Y_0$) [kWh] $\times$ \emph{Gas unit price} ($V_9$) [£/kWh]
\end{itemize}

Capturing this higher-order feedback loop effect is beyond the scope of this analysis. Instead, we focus on identifying the direct \emph{physical} effect of the efficiency intervention ($X \rightarrow Y_0^t$) and the subsequent \emph{behavioural} adjustment mediated by changes in energy cost and burden for the household ($X \rightarrow ...\rightarrow Y_0^{t+\Delta t}$). 

By deliberately omitting second-order feedback effects, we can simplify the graphical representation of the response variable to energy efficiency interventions such as external wall insulation. Specifically, we replace the temporally directed cyclic chain (shown in Figure \ref{fig:main_graphs}-b and Eq. (\ref{behavioural_chain})) with an equivalent acyclic graph involving two atemporal paths, both starting at $X$ and eventually reaching the response variable $Y_0$ from separate ways:

\begin{equation}   
\begin{split}
\label{two_chains}
    X & \rightarrow Y_0  \\
    X & \rightarrow V_1 \rightarrow W \rightarrow Y_0 
\end{split}
\end{equation}

The first pathway represents the \emph{physical} causal mechanism through which an energy efficiency intervention affects gas consumption for space heating. In the specific case of wall insulation, this direct effect operates by reducing the thermal transmittance (U-value) of the building envelope, thereby decreasing the rate of heat loss. The term \emph{direct effect} refers here to the expected change in gas consumption, $Y_0$, resulting from changing the state of wall insulation, $X$, from \emph{false} (no insulation) to \emph{true} (insulation installed), while holding constant the mediating variables ---in this case, gas cost and energy burden--- at the levels they would have attained under the external intervention. This notion of a direct effect corresponds to what Robins and Greenland termed the \emph{pure} direct effect \cite{robins1992identifiability} and what Pearl later formalised as the \emph{natural direct effect} within the framework of structural causal models \cite{pearl2001direct}.

In contrast, the second pathway in Eqs. (\ref{two_chains}) captures the behavioural mechanism through which the intervention $X$ affects gas consumption $Y_0$, mediated by changes in gas costs and energy burden. This behavioural pathway is obtained by removing the mediating variable $Y_0^t$ from Eq. (\ref{behavioural_chain}) and introducing an edge from the intervention $X$ pointing directly to gas costs $V_1$ (Figure \ref{fig:main_graphs}-c). This modification transforms the initial cyclic graph into an equivalent acyclic graph, that is a methodological prerequisite for causal graphical modelling \cite{pearl2009causality}.

\subsection{Accounting for confounders}
\label{sec:acc_confounders}
The core acyclic graph structure, shown in Figure \ref{fig:main_graphs}-c, is further extended by introducing additional variables which may act as potential confounders of the relationship between the intervention ($X$) and the outcome ($Y_0$). A confounder is a variable that influences both the treatment and the outcome, hence failing to account for such variables (e.g. via statistical adjustment) can result in biased estimates of causal effects. 

To mitigate this risk, we expanded the graphical model to include variables that are theoretically expected to confound the primary causal pathways. Importantly, the selection of these variables was guided by theoretical (expert knowledge) understanding of the problem at hand, rather than by data-driven criteria such as statistical association or availability of variables' data in the first place. This approach aligns with best practices in causal DAG construction \cite{tennant2021use, laubach2021biologist, poppe2025develop}, which emphasise the primacy of causal assumptions and substantive knowledge over empirical correlations when specifying the model structure. A detailed description of such causal assumptions is reported in section \ref{Axiomatic_causes} of supplementary material. By prioritising a theory-driven model specification, we aim to ensure that the set of conditioning variables used in the subsequent analysis appropriately blocks non-causal backdoor paths, thereby enabling valid identification of the causal effect of interest. 

This theory-first strategy also addresses a core limitation of discovering causal relations exclusively from data \cite{glymour2019review}, namely: the possibility of fitting multiple DAGs encoding the same set of conditional independencies among observed variables. These DAGs form what is known as a \emph{Markov equivalence class}, i.e. a set of directed acyclic graphs that are statistically indistinguishable based solely on the full joint probability distribution they factorise into (via Eq. (\ref{Bayesian_factorisation})), even though they may imply fundamentally different causal mechanisms (arrow orientations). In the absence of prior causal assumptions or experimental data to rely on, it is therefore not possible to identify the true causal graph within this equivalence class, a point that echoes Cartwright’s dictum \emph{“no causes in, no causes out”} \cite{cartwright1994nocause}, highlighting that causal knowledge cannot emerge from statistical information alone.

By leveraging substantive, theory-driven knowledge of the data-generating process under study, we define a credible DAG from within the equivalence class, which conforms to the statistical properties of the data and also reflects defensible causal relationships grounded in domain expertise.

A key consequence of including variables in the causal graph regardless of their data availability is the appearance of unmeasured variables, denoted $U_i$, which represent latent confounders. Unlike fully observed models, i.e. where all probabilistic dependencies among variables are explicitly captured, the presence of unobserved common-cause confounders can complicate the identification of causal effects because such latent variables cannot be conditioned on (as we have no observational data for them) and therefore identification of unbiased causal effects is not always guaranteed in such settings. We deal with this problem in section \ref{sec_identifiability}.

The final set of variables in the model consists of both observed and latent components. The observed set is given by $\boldsymbol{V}=\{X, Y_0, W, V_0,...,V_{10}\}$, while the unobserved (latent) set is represented as $\boldsymbol{U}=\{U_0,...,U_5\}$. A detailed description of all observed variables is provided in section \ref{subsec_variables_descr} of the supplementary material. The final, fully specified graph, incorporating both observed and latent variables, is shown in Figure \ref{fig:main_graphs}-a.

\subsection{Modularity assumption}
\label{Sec:modularity}
As for Bayesian networks, in a causal graphical model each observed variable $V_i \in\boldsymbol{V}$ is associated with a conditional probability distribution, denoted as $P(V_i=v_i \mid \boldsymbol{Pa}(V_i))_\mathcal{G}$, which describes the probability of $V_i=v_i$ given its parent variables $\boldsymbol{Pa}(\cdot)$ in the graph $\mathcal{G}$ encoding the underlying causal relationships.

A fundamental principle governing causal graphical models is the \emph{causal Markov condition} (also known as modularity assumption), stating that, conditional on its direct parents in the causal graph, each variable is statistically independent of its non-descendants, meaning that once the immediate causal influences of a variable are accounted for (via conditioning), no additional information from its ancestors or other non-descendant nodes contributes to its belief probability. For example, consider a causal chain $X \rightarrow Y \rightarrow Z$, where $X$ denotes the amount of time a student spends studying, $Y$ their resulting exam score, and $Z$ the likelihood of receiving a scholarship. In this case, once we observe the exam score $Y$, knowing how many hours the student spent studying, $X$, provides no further information about their scholarship chances $Z$ (formally: $P(z \mid y, x) = P(z \mid y)$). The effect of $X$ on $Z$ is entirely mediated through $Y$, and the Markov condition implies that $Z \indep X \mid Y$.

As a direct consequence of this assumption, the joint probability distribution over the full set of observed variables $\boldsymbol{V}$ can be expressed as a product of \emph{factors} representing local conditional probability distributions, each corresponding to a variable conditional on its parents in $\mathcal{G}$. This factorisation follows the structural dependencies encoded in the causal graph and is formally given by \cite{koller2009probabilistic}:

\begin{equation}
    \label{Bayesian_factorisation}
    P(v_1,...,v_n)_{\mathcal{G}}= \prod_{i=1}^{n} P(v_i \mid \boldsymbol{Pa}(V_i))_\mathcal{G}
\end{equation}

\subsection{Identifiability}
\label{sec_identifiability}
In this graphical setting, inferring causal effects from observational data requires to establish whether such inferences are theoretically permissible under the assumed causal structure. In particular, if all relevant variables are observed ---what is known as a \emph{Markovian model}--- then post-intervention distributions such as $P(Y \mid do(X=x))$, representing the effect on $Y$ of an external intervention setting $X=x$, can always be derived using a truncated version \cite{pearl2009causality} of the standard Bayesian factorisation Eq. (\ref{Bayesian_factorisation}). Such procedure however relies on the requirement that all potential confounding variables are measurable which is rarely the case in most realistic settings (and our case makes no exception). This more general scenario involving unmeasured confounders $U_i$ in the graph, and referred to as a \emph{semi-Markovian model}, complicates causal effect identification, as identifiability is not always guaranteed in such setting. 

To address this, we employ Pearl’s \emph{do-calculus} \cite{pearl1995causal}, which provides a formal procedure to assesses whether a causal effect can be identified from observational (non-interventional) data, given a specified semi-Markovian causal graph. If identifiability is established, do-calculus yields an explicit formula for estimating the causal effect using only observed quantities, which is reported in the following section.

\subsection{Effects estimation}\label{sec:effect_estimation}

The covariate-specific causal effect of external walls insulation $(X)$ on the response variable gas consumption for space heating $(Y_0)$ in a subset of the sample population for which energy burden $(W)$ achieves a specific value $w$ after the intervention, it is estimated via the following formula:

\begin{equation}
    \label{Eq: covariate_y}
    P( y_0 \mid do(x), w) = \sum_{V_7} {\left[ P\left(y_0 \mid x, w, v_7\right) \left(\frac{\sum_{V_2}{P\left(v_7, w \mid x, v_2 \right) P\left(v_2\right)}}{\sum_{V_2}{P\left(w \mid x, v_2 \right) P\left(v_2\right)}}\right) \right]}
\end{equation}

A formal proof of the estimator formula in Eq. (\ref{Eq: covariate_y}) based on do-calculus computations is provided in the supplementary material section \ref{sec_covar_specific_eff}. 

Eq. (\ref{Eq: covariate_y}) expresses the probability of observing a specific level of gas consumption $Y_0$ in the subgroup defined by energy burden $W=w$, under the intervention $do(X=x)$. The summation over $V_7$ averages over all possible household income levels. The fraction in round brackets acts as a weighting term: it determines how much each income group $v_7$ should contribute when combining the conditional probabilities $P(y_0 \mid x, w, v_7)$. 
These weights are obtained by marginalising over tenancy type ($V_2$). In the causal graph, tenancy is a pre-treatment covariate that mediates the effect of household income ($V_7$) on the probability of receiving insulation ($X$). Statistically, this re-weights the observed income distribution so that it reflects the population structure once insulation has been intervened upon. 
Intuitively, Eq.(\ref{Eq: covariate_y}) combines two pieces of information:  
(i) the conditional relationship between energy use $Y_0$ and the observed attributes $(X, W, V_7)$, and  
(ii) the weighted representation of how frequently each income group $v_7$ occurs in the post-intervention population with energy burden $W=w$, obtained by appropriately averaging over tenancy types.  
Together, this ensures that the resulting estimate corresponds to the distribution of $Y_0$ that would be observed if wall insulation were intervened upon in the real population.

The causal effect of external walls insulation $(X)$ on the response variable $(Y_0)$ across the entire sample population, $P(y_0 \mid do(x))$, can then be derived by applying the law of total probability:

\begin{equation}
    \label{Eq: ce_y}
    P(y_0 \mid do(x)) = \sum_{w} P(y_0 \mid do(x), w) P(w \mid do(x))
\end{equation}
where the first \emph{factor} on the r.h.s. expressed in terms of observational quantities is obtained via Eq. (\ref{Eq: covariate_y}) whilst the second \emph{factor} on the r.h.s., $P(w \mid do(x))$, is derived from observational probability distributions as follows:

\begin{equation}
    \label{Eq: w_do_x}
    P(w \mid do(x)) = \sum_{V_2} P(w \mid x, v_2) P(v_2)
\end{equation}
A formal proof of Eq. (\ref{Eq: w_do_x}) is provided the in supplementary material (Eq. (\ref{Eq: SM_l})). 
Eq. (\ref{Eq: ce_y}) aggregates the subgroup-specific distributions from Eq. (\ref{Eq: covariate_y}) into the overall population-level post-intervention distribution. The term $P(y_0 \mid do(x), w)$ captures the effect of the intervention within the subgroup defined by burden level $w$, while $P(w \mid do(x))$ represents how frequently that subgroup occurs in the population once insulation has been interveend upon. Multiplying and summing out across all values of $w$ therefore combines the within-group causal effects into a single population-wide effect. Formally, this is an application of the law of total probability, but intuitively it means that the overall effect is obtained as a weighted average of subgroup effects, where the weights correspond to the prevalence of each subgroup after the intervention.

Notably, the unbiased population-wide causal effect $P(y_0 \mid do(x))$ can been obtained ---as alternative to Eq. (\ref{Eq: ce_y})--- using the \emph{back-door criterion} \cite{pearl2009causality}, which involves identifying a set of adjustment variables satisfying the following two conditions:

\begin{itemize}
    \item [-] No variable in the adjustment set is a descendant of $X$
    \item [-] The adjustment set blocks every back-door path between $X$ and $Y_0$
\end{itemize}
A back-door path is any path from $X$ to $Y_0$ in the causal graph $\mathcal{G}$ that includes an arrow pointing into $X$, thus representing a non-causal path that may introduce confounding bias. For example, variable $V_2$ satisfies both conditions and thus qualifies as a valid (and sufficient) adjustment set to compute $P(y_0 \mid do(x))$ (see Figure \ref{fig:main_graphs}). 

However, our goal extends to estimating the covariate-specific causal effect for subpopulations defined by a fixed energy burden value $W=w$. This requires that all back-door paths from $X$ to $Y_0$ remain blocked after including $W$ in the conditioning set \cite{pearl2016causal}. Unfortunately, this is not feasible under the back-door criterion because $W$ is a descendant of $X$, and including it in the adjustment set would violate the first of the two back-door conditions stated above. Indeed, cautionary guidance in the literature has long warned against conditioning on post-treatment variables due to risks of introducing collider bias \cite{montgomery2018conditioning}.

To address this, we resort to \emph{do-calculus}, which enables principled reasoning about such cases. Contrary to general intuition, the do-calculus framework establishes that conditioning on a descendant of $X$ does not necessarily bias the estimate of the causal effect \cite{pearl2015conditioning} as long as the independency condition stated by its three rules (see supplementary material section \ref{do_calc_rules}) are being met. We derive the custom identification formula presented in Eq. (\ref{Eq: covariate_y}) accordingly. This enables unbiased estimation of the $W$-specific causal effect of $X$ on $Y_0$, despite $W$ being a post-treatment variable.

Interestingly, the covariate-specific estimator formula so obtained (Eq. (\ref{Eq: covariate_y})) requires observing only four variables $\{X, W, V_2, V_7 \}$ in order to compute the post-intervention distribution $P( y_0 \mid do(x), w)$. This set forms a \emph{sufficient} adjustment set, as it blocks all non-causal paths whilst conditioning on a collider node ($W$).  

All of the observational conditional probabilities on the r.h.s. of Eqs. (\ref{Eq: covariate_y}), and r.h.s. of Eq. (\ref{Eq: w_do_x}), are computed from the Bayesian network factorisation (i.e. Eq. (\ref{Bayesian_factorisation})) of the joint $P(\boldsymbol{V})$ using Variable Elimination (VE), a recursive algorithm to perform \emph{exact} inference in probabilistic graphical models \cite{zhang1994simple}. VE simplifies the computation of marginal probabilities by iteratively eliminating variables and combining \emph{factors} until only the queried marginal or conditional probability remains. Unlike \emph{approximate} inference methods such as those based on Monte Carlo sampling, VE yields exact results that are  not subject to approximation errors arising from sampling variability or convergence criteria.

\subsection{Average and conditional treatment effects}
\label{sec:ATEs_comp}
Post-interventional probability distributions obtained via Eqs. (\ref{Eq: covariate_y}-\ref{Eq: ce_y}) are then used to compute the corresponding expectation (sample's mean) value of gas consumption $Y_0$ given the external intervention on wall insulation: $X=true$ ($=false$):

\begin{equation}
\label{eq:expectation}
\begin{split}
    & \mathbb{E}(Y_0 \mid do(x)) = \sum^n y_0  P(y_0 \mid do(x)) \\
    & \mathbb{E}(Y_0 \mid do(x), w) = \sum^n y_0  P(y_0 \mid do(x), w) \\
\end{split}
\end{equation}

with $n$ being the number of unique (discretised) assignment values $y_0$ for the real-valued variable $Y_0$.
Eqs. (\ref{eq:expectation}) shows how post-intervention probability distributions (from Eqs. (\ref{Eq: covariate_y}–\ref{Eq: ce_y})) are converted into expected values of gas consumption. These expected values are the building blocks of treatment effect estimation, since they provide the average outcome we would see under each intervention scenario.

Accordingly, both the average treatment effect (ATE), representing the treatment’s impact on gas consumption across the entire sample population, and the conditional average treatment effect (CATE), representing the treatment’s impact within subgroups of the population characterised by a specific level of energy burden $W=w$, can both be expressed as follows: 

\begin{equation}
\label{EQ:ATE_CATE_G}
\begin{split}
    & ATE_{\mathcal{G}} = \mathbb{E}(Y_0 \mid do(X=true))_{\mathcal{G}} - \mathbb{E}(Y_0 \mid do(X=false))_{\mathcal{G}} \\
    & CATE_{\mathcal{G}} = \mathbb{E}(Y_0 \mid do(X=true), w)_{\mathcal{G}} - \mathbb{E}(Y_0 \mid do(X=false), w)_{\mathcal{G}} \\
\end{split}
\end{equation}

Specifically, Eqs. (\ref{EQ:ATE_CATE_G}) formalise the average treatment effect and conditional average treatment effect as differences between the expected outcomes with and without insulation. The ATE captures the effect across the full population, while the CATE isolates the effect for households with a specific level of energy burden.

Of note, the subscript $\mathcal{G}$ in Eq. (\ref{EQ:ATE_CATE_G}) indicates that treatment effects are inferred based on the factorisation of $P(\boldsymbol{V})$ compatible with the causal structure encoded in the graph $\mathcal{G}$ ---i.e. the factorisation stated in Eq (\ref{Bayesian_factorisation}). This means that both the physical causal effect of the wall insulation intervention on gas consumption ($X\rightarrow Y_0$) and the behavioural effect mediated by energy burden ($X\rightarrow V_1 \rightarrow W \rightarrow Y_0$) are taken into account in the estimation of $ATE_{\mathcal{G}}$ and $CATE_{\mathcal{G}}$.
A step-by-step illustrative example of ATE and CATE calculation is provided in the supplementary material section \ref{ATECATE_dummy_exemple}.

\subsection{Handling selection bias of gas cost data}

Household gas cost data for space heating, as reported in the EHS-FP dataset ($V_1$) are derived using the BRE domestic energy model (BREDEM), a rule-based building-physics engineering model developed by the Building Research Establishment \cite{ henderson2012bredem}. BREDEM estimates heating requirements under standardised assumption: households are expected to maintain demand temperatures of 21\textdegree C in living rooms and 18\textdegree C elsewhere during heating session \cite{FulePoverty_methodology}, with heating demand calculated based on several factors such us dwelling characteristics, heating system type, local weather, and household composition \cite{FulePoverty_methodology}. The methodology also accounts for demographic variation (e.g. longer heating sessions for elderly or households with young children, reduced heating of unused rooms in under-occupied dwellings).

As a result, reported gas costs reflect simulated rather than observed gas expenditures. These values are accurate insofar households heat in line with the BREDEM model assumptions. In practice, financially constrained households may underheat to reduce bills, meaning BREDEM-based costs may likely overestimate actual spending for high-burden households ---an issue central to the present study, as we test the hypothesis that behavioural factors may lead to deviations from adequate heating regimes. This raises the question of whether such behavioural effects can be reliably detected by training our causal ML model using EHS-FP data for space heating cost. 

To address this we reframe the issue as a problem of \emph{sample selection bias} \cite{heckman1979sample, lee2005training}. The bias arises because households that maintain adequate heating despite high energy burden are more likely to be represented in the BREDEM-based dataset, while those who underheat are systematically under-represented.
Consequently, the EHS-FP dataset constitutes a biased sample of the target population. It can be interpreted as if (actual) gas cost data were collected more frequently from household units engaging in adequate heating while systematically excluding units engaging in financially driven energy saving behaviour. 

To handle this inherent bias in the training dataset, we follow on works from Bareinboim et al. \cite{bareinboim2015recovering, bareinboim2016causal, bareinboim2022recovering} which provide causal graphical criteria to test whether it is possible to recover population-wide observational (or post-interventional) distributions from samples collected under preferential selection. In our causal graphical model, we introduce a binary selection variable $S$, where $S=1$ denotes household units were included in the sample dataset, and $S=0$ indicates otherwise. To capture the underlying causal mechanism biasing the data-gathering process, we introduce an arrowed edge from $W$ to $S$ in the graph $\mathcal{G}$, thus obtaining an augmented graph $\mathcal{G_S}$ in which $W\rightarrow S$ conveys the causal effect of energy burden on dataset inclusion/exclusion for the unit. This reflects the idea that household units experiencing a high energy burden ($W=high$)  are more likely to adopt heating behaviours that deviate from the assumptions made in the energy simulation model, thus leading to their exclusion from the collected sample. 

Our goal is to obtain the post-intervention probability distribution of $Y_0$ representative of the target population, i.e. $P(y_0 \mid do(x), w)_{\mathcal{G_S}}$, but since the sample was collected under selection bias, only the distribution conditional on $S=1$ is accessible for use, i.e. $P(y_0 \mid do(x), w, S=1)_{\mathcal{G_S}}$. As such, the requirement under which the population-wide causal effect is recoverable from the biased sample essentially entails for these two probability distributions to be equivalent. This can be verified by leveraging on rule 1 of Pearl's do-calculus \cite{pearl2009causality} (see section \ref{do_calc_rules} in supplementary material):  

\begin{equation}
    \label{Eq: debias}
    P(y_0 \mid do(x), w)_{\mathcal{G_S}} = P(y_0 \mid do(x), w, S=1)_{\mathcal{G_S}} \mbox{ \ if: \ } \left(Y_0 \indep S \mid X, W \right)_{\mathcal{G_{S,\overline{X}}}}
\end{equation}

Compliance to the conditional independency requirement stated in Eq. (\ref{Eq: debias}) can be checked graphically via the \emph{d-separation} criterion \cite{geiger1990d}, by inspecting the mutilated augmented graph $\mathcal{G_{S,\overline{X}}}$, that is $\mathcal{G_{S}}$ after removing all edges pointing to $X$ (see Figure \ref{fig:subgraphs} in supplementary material). It suffice here to note that by conditioning on $W$ we block any statistical association between nodes $S$ and $Y_0$ in $\mathcal{G_{S,\overline{X}}}$, thus rendering them conditionally independent given $X$ and $W$.

\subsection{Methodological limitations}
Our estimates rely on the assumed DAG, which is grounded in substantive theory and empirical evidence (section \ref{subsec_direct_causal_relations} in supplementary material). However the choice of its structure remains a modelling assumption. Misspecification of key causal pathways (for example, omitting a relevant confounder or incorrectly orienting an arrow) could affect the estimated causal effects. We have explicitly discussed this dependence in section \ref{sec:acc_confounders} and acknowledge it as a limitation of the study.
Importantly, the adjustment set used in our analysis is sufficient under the specified DAG, meaning it blocks all non-causal backdoor paths to isolate the effect of $X$ on $Y_0$. Nevertheless, readers should interpret the results in the context of the assumptions encoded in the DAG. This limitation is not unique to causal graphical models; all causal inference methods (e.g. including matching and difference-in-differences) crucially rely on untestable assumptions about the underlying data-generating process.\footnote{Matching methods rely on the Conditional Independence Assumption, implying that all confounding variables have been included and controlled for in the model. In the language of causal graphical models, this is equivalent to finding a \emph{sufficient} adjustment set of covariates that blocks all non-causal paths between treatment and outcome (see section \ref{sec:effect_estimation} and Figure \ref{fig:main_graphs}). Difference-in-Differences (DiD) methods relies on the Parallel Trends Assumption, that is, the treated and control groups would have followed the same outcome trend without the intervention.} Our theory-driven DAG specification increases transparency (and hence defensibility) of these assumptions.

A further limitation concerns the use of BREDEM-simulated rather than observed gas expenditures. Simulated costs provide a consistent benchmark but may overstate actual spending for high-burden households who underheat, leading to sample selection bias. Our approach (following Bareinboim et al. \cite{bareinboim2015recovering, bareinboim2016causal, bareinboim2022recovering}) addresses this issue by using causal graphical criteria to establish when population-level effects can still be recovered despite preferential sampling. This provides a theoretical guarantee that bias remains low under the stated assumptions. However, without observed (metered) gas expenditures of the same household units, the extent of any residual bias cannot be empirically quantified. Importantly, using metered gas expenditure data from other datasets would not automatically resolve this issue, because these data would come from households different from those in the EHS-FP sample. Without a shared unique identifier (such as a household ID) across the datasets, a standard data linkage or merge is unfeasible. In a statistical modelling context, the two populations may also lack sufficient common support in their overall characteristic distributions, which would prevent the reliable comparison of energy use patterns. Future work could address this limitation through data fusion methods, provided that either uniquely linkable or otherwise strongly comparable household-level data are available.

\section{Results}\label{sec_results}

Our analysis aims to estimate the causal impact of energy efficiency interventions on energy use, which in turn influences energy costs and, consequently, the economic burden on households. To do so, we focus on a specific intervention: the installation of external wall insulation, which we denote as the treatment variable $X$ in our causal model (Methods section \ref{sec_methods}). While other retrofit interventions may significantly reduce energy demand, not every efficiency intervention necessarily leads to cost savings as well. That is the case for example of replacing a gas boiler with an electric heat pump: a typical domestic heat pump delivers between 3 to 5 units of heat for every unit of electricity used \cite{rosenow2022heating}, far outperforming even the most efficient gas boilers, which operate at around 95\% efficiency. However, the higher retail unit price of electricity relative to gas (as of January 2025, $\approx$0.25£/kWh for electricity versus 0.06£/kWh for gas; UK averages) offsets much of this efficiency gain. As a result, households transitioning from gas to heat pumps would observe much lower energy demand but not proportionally lower energy bills, especially in the absence of time-of-use tariffs or government subsidies.
By contrast, interventions such as building fabric upgrades reduce energy required to maintain indoor thermal comfort regardless of the energy vector and heating system technology installed, making them well-suited to evaluate our central hypothesis around fuel burden alleviation. 

For this reason, we selected external wall insulation as our treatment variable ($X$) throughout the analysis. The estimated effects presented in the following subsections are measured in terms of annual gas consumption levels for space heating (in kWh/yr) denoted by the variable $Y_0$.

\begin{figure}[ht]
    \centering
    \includegraphics[width=9.5cm]{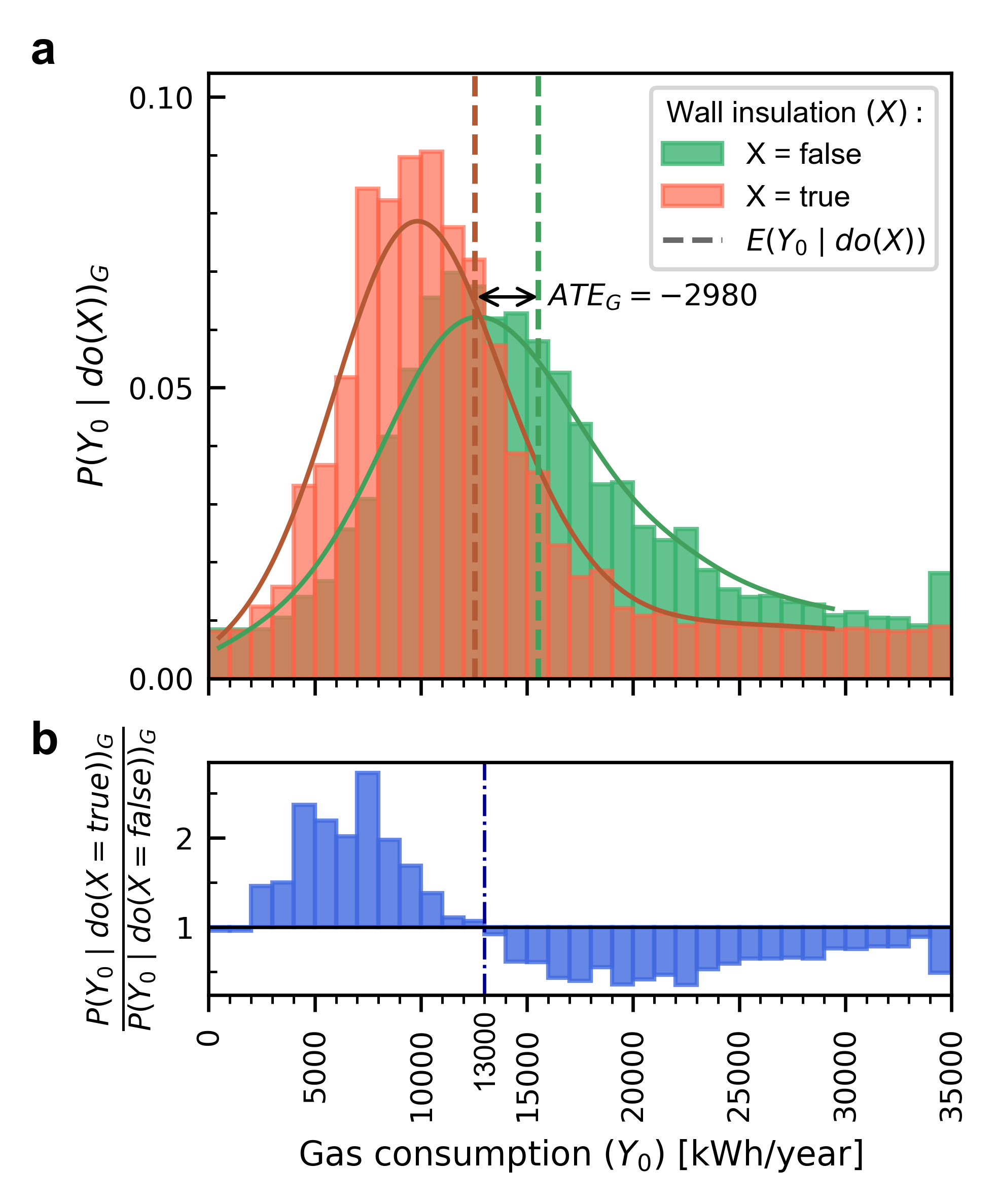}
    \caption{\small Estimated causal effect of walls insulation intervention, $X$, on annual gas consumption for space heating, $Y_0$. \textbf{a}: post-intervention distributions of gas consumption under the two interventional scenarios: $P(Y_0 \mid do(X=true))$ i.e. treatment group, and $P(Y_0 \mid do(X=false))$ i.e. control group. The downward shift in expectation $\mathbb{E}(Y_0 \mid do(X))$ under treatment $X=true$ indicates a clear average reduction in energy use ($\approx-2980$ kWh/year). \textbf{b}: probability ratio, $PR(Y_0)$, plotted across the range of annual gas consumption levels. $PR(Y_0) > 1$ suggests an increased likelihood of remaining in a low-consumption regime post-insulation, while $PR(Y_0) < 1$ indicates a reduced likelihood of persisting in high-consumption states. A clear pattern emerges, with higher-consuming households ($>13,000$ kWh/yr) consistently more likely to reduce their energy use compared to lower-consuming households.}
    \label{fig:ATE_0}
\end{figure}

\subsection{Average Treatment Effect (ATE)}
We begin by reporting the population-wide Average Treatment Effect (ATE) of external wall insulation on annual gas consumption for space heating. Here, the ATE is defined as the difference in expected outcomes $\mathbb{E}(Y_0 \mid do(X))$ under two scenarios: one in which all dwellings in the population sample are externally insulated ($X = \text{true}$), and another in which none receives the intervention treatment ($X = \text{false}$) (see section \ref{sec:ATEs_comp}). 

Figure \ref{fig:ATE_0}-a shows the estimated post-intervention distributions $P(Y_0 \mid do(X))$ under both treatment conditions, along with their corresponding expectations $\mathbb{E}(Y_0 \mid do(X))$. 

It is important here to distinguish the causal quantity $P(Y_0 \mid do(X))$ from its observational counterpart $P(Y_0 \mid X)$. While $P(Y_0 \mid X)$ tells us how gas consumption differs between dwellings that happen to be insulated and those that are not, it does so within the structure of the existing system, where insulation status may be influenced by other factors such as income, property type, occupant behaviour, etcetera (section \ref{sec:acc_confounders}). 
By contrast, $P(Y_0 \mid do(X))$ represents a hypothetical intervention, capturing what the distribution of gas consumption would look like if we were to externally impose insulation (or lack thereof) across all dwellings, breaking any natural dependencies that normally determine who receives the intervention. In this sense, conditioning on $X$ reflects correlation within the current world, while intervening with $do(X)$ imagines an altered world in which we directly set the value of $X$ and observe the resulting changes in $Y_0$. This distinction is foundational to causal inference, and it underpins our interpretation of the average treatment effect reported here \cite{pearl2009causality}.

According to our findings, external wall insulation results in a population-wide average reduction of approximately 2980 kWh/year in gas consumption for space heating, from a baseline average annual consumption of $\approx$15,400 kWh/year. This corresponds to a mean decrease of about 19\% relative to the untreated group.

The estimated 19\% reduction in gas consumption is substantially higher than the effect sizes reported in previous causal inference (difference-in-differences) evaluations of domestic insulation in England. Peñasco and Anadón \cite{penasco2023assessing} find that cavity wall insulation reduces annual gas use by approximately 6.9\% in the short term, with the effect tapering off over a four-year period. Similarly, Adan and Fuerst \cite{adan2016energy} report average reductions of around 10.5\%. While these figures provide benchmarks for comparison, they refer specifically to cavity wall insulation which typically applies to relatively modern housing, with better thermal performance at baseline. 
By contrast, our analysis does not distinguish between cavity and solid wall types, as our focus is on the overall effect of insulation regardless of wall construction. This approach includes a broader sample, capturing more older, less energy-efficient homes, particularly those with solid walls, where insulation typically leads to greater gas savings.
Additionally, both studies rely on data from the LEED dataset, which is constructed from observed participants in government retrofit schemes. As such, the sample may be subject to self-selection bias, since households opting into such schemes may differ systematically from the general population. 
In contrast, our use of a randomly selected (EHS) dwelling stock sample avoids this particular source of bias. Finally, it remains unclear in the literature whether reported percentage changes in metered gas consumption, based on the LEED dataset, refer specifically to space heating or to total household gas use, i.e. including other end-uses such as cooking and water heating. This ambiguity may contribute to discrepancies in reported effect sizes, especially given that insulation interventions only affect the share of gas consumption for the specific use of space heating.

Beyond average effects, the distributional impact of external wall insulation can be further explored through analysis of the probability ratio $PR(Y_0)$, defined as the ratio of post-intervention outcome probabilities under treatment and control: $PR(Y_0) = \frac{P(Y_0 \mid do(X=true))}{P(Y_0 \mid do(X=false))}$. As shown in Figure \ref{fig:ATE_0}-b, this ratio captures how the likelihood of a given level of gas consumption changes under the interventional scenario compared to the baseline (i.e. untreated) scenario. The analysis reveals a clear threshold effect at around 13 MWh/year. 
For households with baseline gas consumption below this value, the probability ratio consistently exceeds 1, indicating that external wall insulation increases the likelihood of remaining in a low-consumption regime. In contrast, for households with higher baseline usage (above 13 MWh/year), $PR(Y_0) < 1$, thus implying a reduced likelihood of persisting in high-consumption states post-intervention. This indicates a general redistribution of the probability mass, shifting toward lower levels of gas consumption under treatment. It also indicates a clear distributional effect trend of the intervention across the sample population, in that high-consuming households are systematically more likely to reduce their gas consumption levels post-treatment compared to those with a pre-treatment consumption level below the 13 MWh/year threshold.

\begin{figure}[ht]
    \centering
    \includegraphics[width=12.9cm]{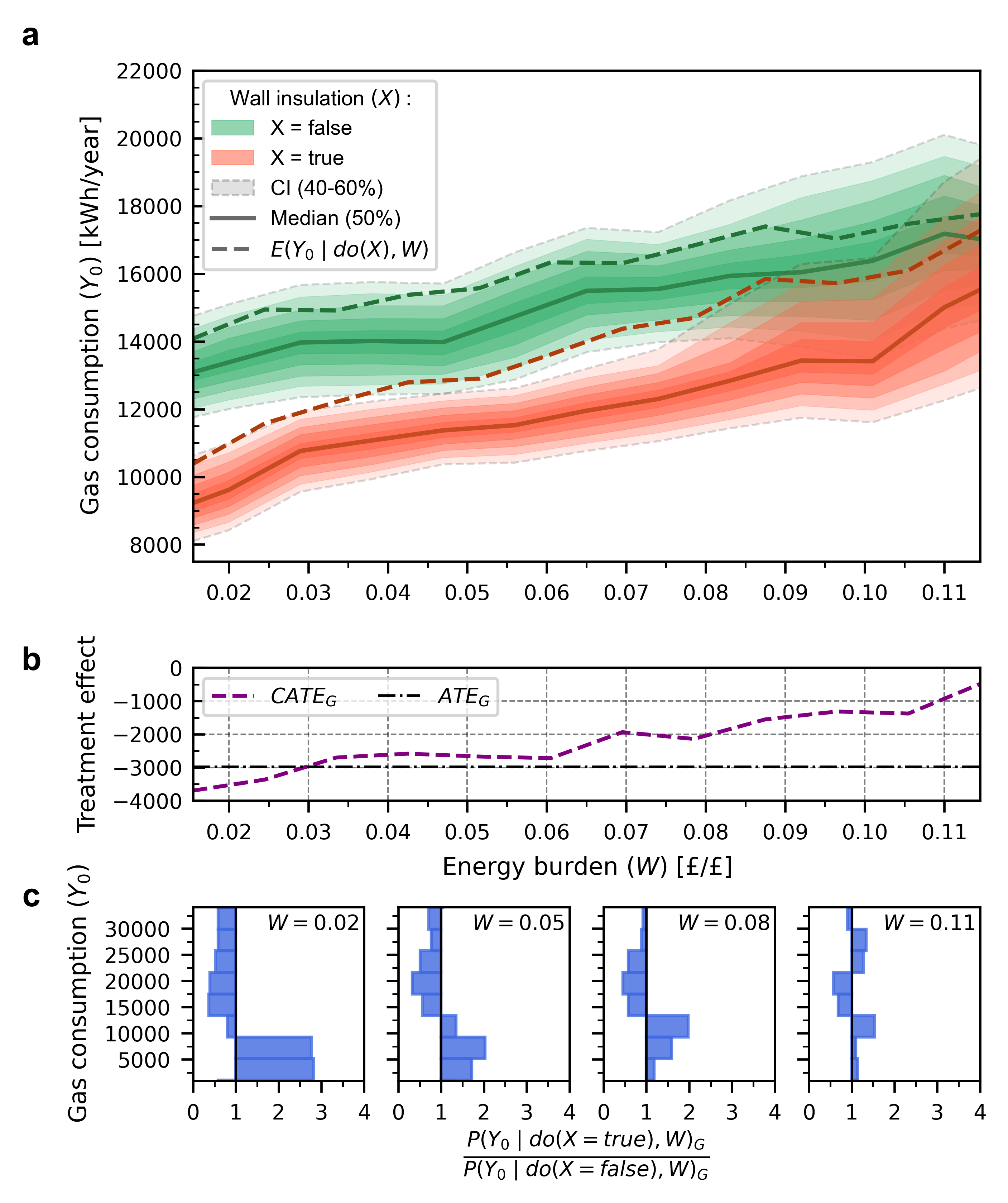}
    \caption{\small Conditional average treatment effects (CATE) of external wall insulation on annual gas consumption by energy burden subgroups. \textbf{a}: estimated post-intervention outcome distributions $P(Y_0 \mid do(X), W)$ by 
    energy burden level $W$, defined as the share of income spent on energy. Confidence intervals (40–60\%), medians, and conditional expectations $\mathbb{E}(Y_0 \mid do(X), W)$ are reported for both treated and untreated scenarios. \textbf{b}: plots of the Conditional Average Treatment Effect (CATE) as a function of $W$, revealing a declining trend in treatment efficacy with increasing energy burden: households spending 2\% ($\pm 0.5\%$) of income on energy see reductions of $\approx 3720$ kWh/year ($\approx –26\%$), whereas the most burdened ($W>11\%$) only achieve $\approx$500 kWh/year ($\approx –3\%$). \textbf{c}: probability ratios $PR(Y_0)$ of population subgroups experiencing different levels of energy burden, showing that low-burden households with medium-to-high baseline gas consumption are consistently more likely to reduce demand post-intervention ($PR(Y_0) < 1$). This pattern disappears at higher burden levels (e.g. $W=0.11\pm 0.5\%$), where all households show limited post-treatment change, regardless of initial gas consumption.}
    \label{fig:CATE_0}
\end{figure}

\subsection{Conditional Average Treatment Effect (CATE)}
\label{subsec:cate_results}

While the ATE provides a summary statistic of the effect of external wall insulation across the entire population, it does not reveal how the intervention impacts households facing different levels of energy vulnerability, an aspect that is instrumental to the main hypothesis being tested. To address this, we estimate the Conditional Average Treatment Effect (CATE), defined as the difference in expectation outcomes $\mathbb{E}(Y_0 \mid do(X), W)$  for the interventional scenario $X=true$ and baseline scenario $X=false$, now conditional on observing specific levels of energy burden (see section \ref{sec:ATEs_comp}). This latter is denoted by $W$, and defined as the ratio of overall energy costs to household income, a widely used metric to capture vulnerability to fuel poverty. 

Figure \ref{fig:CATE_0}-a presents the estimated post-intervention outcome distributions $P(Y_0 \mid do(X), W)$ for households grouped into discrete energy burden strata, ranging from 2\% ($\pm0.5\%$) up to 11\%. For each stratum, confidence intervals (40–60\%) are shown for both treatment conditions, together with the corresponding medians and conditional expectations $\mathbb{E}(Y_0 \mid do(X), W)$. We observe that average gas consumption rises with energy burden, irrespective of treatment status. However, as revealed by the distance between treated and untreated distributions, the effect of insulation declines steadily as energy burden increases. This results in greater overlap between the two distributions for high-burden groups, signalling a reduced intervention impact for these households.

This heterogeneity is more clearly illustrated in Figure \ref{fig:CATE_0}-b, which shows CATE values across energy burden levels. Households with relatively low burden (1.5–2.5\% of income spent on energy) experience the largest treatment effect, with an average gas reduction of approximately 3720 kWh/year ($\approx–26$\% relative to pre-intervention level). This benefit decreases progressively, reaching only $\approx$500 kWh/year ($\approx–3\%$) for households in the highest burden category ($W>11\%$).

Furthermore, analysis of the probability ratio across energy burden subgroups (Figure \ref{fig:CATE_0}-c) provides additional evidence of the uneven distributional effects. Among households with low energy burden ($W = 2\% \pm 0.5\%$), those with mid-to-high baseline gas consumption (i.e., $>9,000$ kWh/yr) are consistently more likely to reduce their demand following the insulation intervention, as indicated by probability ratios $<1$. However, this pattern progressively fades as energy burden increases. At higher burden levels, such as $W = 11\%$ ($\pm 0.5\%$), no clear threshold between high and low gas consumers is observed.

Here, probability ratios flatten and converge toward unity across the board, independently of pre-intervention consumption levels, thus suggesting a reduced differential impact of the intervention. This confirms that for households facing higher energy burden, the likelihood of achieving substantial reductions in gas consumption diminishes regardless of their absolute consumption levels. One likely explanation is that these households are already limiting their energy use to prioritise other essential expenditures (e.g. food, rent, debt repayments). As a result, enhancing the energy efficiency of their dwelling does not lead to further meaningful reductions in gas consumption, since energy demand was already suppressed prior to the intervention, that is strong evidence of prebound effect. Arguably, for the most vulnerable groups, financial constraints (rather than energy inefficiency) is the primary barrier preventing them to lower their carbon footprint further.

\subsection{Discussion}\label{sec_discussion}

Energy efficiency upgrades lead to significant energy reductions for households with low energy burden. For high-burden households, reductions are more modest, suggesting financial constraints and/or behavioural adaptation may limit the estimated carbon savings.

A critical implication of our results is the emergence of a structural trade-off between mitigating climate change and improving household well-being. Energy efficiency upgrades lead to significant energy reductions for households with low energy burden. In contrast, for high-burden households, the intervention yields relatively modest gas consumption reductions. This suggests that financial constraints limit further carbon savings for this group. The explanation lies in behavioural adaptation post-intervention: households reallocate the financial savings from the intervention toward increasing indoor temperatures, thereby alleviating thermal discomfort. While this behavioural shift may limit the net carbon savings predicted by engineering models, if households use the resulting financial savings to increase indoor temperatures, this could imply social co-benefits for occupants, although we do not directly measure such outcomes.

Our analysis did not directly measure the causal effect on indoor temperature levels across the household population pre- and post-intervention, meaning we cannot provide direct empirical evidence that efficiency interventions lead to adequate temperature levels for subgroups that may engage in underheating behaviours.
We acknowledge the possibility that some households might simply engage in wasteful heating behaviours (e.g., leaving windows open while heating) once they can afford it, due to exogenous behavioural factors. While our causal graphical model is specifically designed to account for additional behavioural effects of latent confounders such as educational attainment ($U_2$) or other exogenous variables ($U_5$) affecting both gas consumption and the likelihood of high energy burden, future research could complement this approach by including household-level measurements of indoor temperature or other comfort indicators before and after interventions. Such data would allow empirical validation of whether the estimated behavioural adaptation indeed improves well-being. 
As explicitly visualised in Figure \ref{fig:main_graphs}-a, our model ensures that any statistical association measured between the intervention and the outcome, when mediated via energy burden ($W$), can be interpreted in causal terms. This means our findings represent the isolated causal effect of financial relief resulting from the efficiency intervention ($X \rightarrow V_1 \rightarrow W \rightarrow Y_0$), insofar the specified causal graph \emph{faithfully} captures the data generating process being modelled (Methods section).

Although our analysis is limited to the UK, the underlying behavioural mechanism (whereby energy efficiency interventions enable households with high energy burden to shift from energy deprivation to improved thermal comfort) it is likely to have broader relevance. In particular, similar dynamics may be observed in countries of the Global South, where energy poverty is widespread and housing quality is often poor, with limited access to affordable energy services. As these regions undergo rapid urbanisation and face increasing cooling demand due to climate change, energy efficiency policies should anticipate the same tension between carbon mitigation and the need to ensure liveable indoor environments. Future policy and research efforts should also account for cultural and climatic variability in household energy practices and thermal expectations, which may shape both the direction and magnitude of behavioural rebound effects in diverse contexts.

\subsection{Policy implications}
 
Our findings pose a significant challenge to prevailing methodologies in building energy modelling and policy evaluation. Energy and climate policies are frequently framed with the explicit intent of delivering dual benefits: simultaneously contributing to GHG emission reduction targets and addressing social inequalities, particularly fuel poverty and its associated health impacts \cite{beis2021net, hbs_2021, ofgem_eco}. However, our study reveals a crucial nuance. Academic literature and policy assessments typically project energy efficiency intervention impacts assuming idealised baseline conditions, resulting in uniform energy savings across diverse occupant populations, often through the use of simplified building archetype models  \cite{nayak2023review, shen2024archetype} that do not capture behavioural heterogeneity due to pre-existing energy deprivation. Consequently, conclusions drawn from these methods offer only a partial understanding of actual energy savings and their distribution, potentially leading to overestimations of carbon reductions while significantly undervaluing the crucial social benefits related to improved liveability.

The ``win-win'' narrative that energy efficiency simultaneously delivers substantial carbon emission reductions and improved household well-being requires critical re-evaluation. Our results reveal a structural trade-off in real-world applications of domestic energy efficiency interventions, particularly in the context of fuel poverty. For households with low energy burdens, insulation effectively delivers climate gains. These households are able to reduce gas consumption substantially without compromising indoor comfort, translating into tangible carbon savings aligned with engineering model predictions.

However, for those experiencing high energy burdens, the primary benefit of the intervention shifts from maximising carbon savings to potentially alleviating thermal discomfort and enhancing well-being. These households rationally reallocate financial relief to achieve healthier indoor temperatures, resulting in an increase in energy use aimed at improving comfort, which, while profoundly beneficial socially (e.g., reducing respiratory/cardiovascular illnesses and excess winter mortality \cite{ballesteros2022effects, mei2024relationships, moorcroft2025damp, singh2022estimating}), significantly diminishes the intervention's net carbon mitigation potential for this group. These crucial social welfare benefits are rarely captured in carbon accounting models informing climate policy, thus underestimating the total societal value of energy efficiency interventions.

This nuanced trade-off is particularly pertinent given recent energy price surges (2021-2023) across the UK and Europe, which have pushed more households into higher energy burden strata \cite{DESNZ2023Subnational}. In this evolving landscape, the role of energy efficiency interventions may increasingly be understood as a means to alleviate thermal discomfort and fuel poverty, rather than solely as a tool to drive down GHG emissions. Policies that do not account for these divergent outcomes and the behavioural adaptations they trigger risk misrepresenting their true impact and inadvertently overlooking the complex needs of vulnerable populations \cite{BBC2023}. For example, UK schemes like the Energy Company Obligation (ECO) \cite{ofgem_eco}, which target low-income, fuel-poor households, may miscalculate their carbon contribution if they solely rely on models that assume uniform energy savings. Policymakers could refine such programmes by explicitly considering household energy burden profiles, targeting support to high-burden households with interventions that simultaneously alleviate fuel poverty and account for expected behavioural adaptation.

We advocate for an integrated policy framework that explicitly incorporates social welfare objectives, such as thermal comfort and public health indicators, into carbon accounting models. For instance, carbon accounting frameworks could weight avoided emissions differently for households where energy efficiency upgrades also deliver measurable health and well-being improvements, or include metrics such as reductions in cold-related morbidity and excess winter mortality.
This approach would ensure a holistic assessment of energy efficiency measures, valuing both their climate and crucial equity contributions. Positioning energy efficiency as part of a broader strategy to reduce both emissions and social inequalities will ultimately contribute to a more just and sustainable transition to a low-carbon economy.

\subsection{Directions for future research}
In this study, we focused on income and energy burden as the primary sources of heterogeneity, as these dimensions directly capture distributional concerns central to energy poverty and efficiency policy. We acknowledge that additional factors ---such as dwelling type, geographic region, tenure status, or urban/rural distinctions--- could offer further insights into the differential impacts of insulation interventions.
However, some of these dimensions cannot be fully explored with the current dataset; for instance, urban/rural location and fine-grained geographic identifiers are not available in the publicly accessible version of the English Housing Survey. For other factors, such as dwelling type and tenure, the causal model already incorporates related structural features (e.g., wall type, ownership vs. rental) within the adjustment set, though these are not analysed as separate heterogeneity dimensions.
Extending the heterogeneity analysis to cover all suggested dimensions would require additional data or an alternative study design, which lies beyond the scope of the present work. We highlight this as a valuable direction for future research.

\section{Refutation}
\label{subsec:refutation}
Causal effects estimation relies on critical assumptions regarding the data-generating process (see Methods section \ref{sec_methods}). As a result, any violation of these assumptions can lead to significant inaccuracies in the estimated effects. In contrast to ML models aimed at purely predictive tasks, where cross-validation offers a comprehensive measure of model performance, causal inference lacks a universally applicable validation method. To tackle this issue, we draw on recent work \cite{sharma2021dowhy} grounded in the philosophical principle of \emph{falsifiability} \cite{popper1934logic} ---positing that scientific claims cannot be conclusively proven, but gain credibility by resisting empirical attempts at refutation--- we conduct refutation checks to test the reliability of our causal estimates. Aim of these tests is to check whether the estimated treatment effect conforms to the theoretical expectation required for the estimator to produce valid causal inferences. Such expectations will depend on the specific test, as explained in each of the following subsections.

\subsection{Placebo treatment test}
\label{section_placebo_tretment_test}
First, we perform a placebo treatment test \cite{eggers2024placebo} by randomly permuting the values of the treatment variable $X$ in the training dataset. This involves assigning to each household unit a treatment value $x$ drawn at random from the original dataset, effectively reallocating treatment values across units. This process replaces the original treatment variable $X$ with a randomised version, $X_{\text{placebo}}$, thereby `breaking' any causal association between $X$ and its children variables $\textbf{ch}(X)$, while preserving its marginal probability distribution: 

\begin{equation}
\label{eq:placebo_1}
\begin{split}
    & P(X_{\text{placebo}}) = P(X) \\
    & P(\textbf{ch}(X_{\text{placebo}}) \mid X_{\text{placebo}}) \neq P(\textbf{ch}(X) \mid X) \\
\end{split}
\end{equation}

We then evaluate whether the causal estimator (Eqs. (\ref{Eq: covariate_y}-\ref{EQ:ATE_CATE_G})) detects any significant association under this placebo setting. Under valid causal assumptions, the theoretical expectation is that the estimator should yield a null average treatment effect for the randomised treatment: $ATE_{\text{placebo}} \approx 0$. While this test does not directly assess the estimated treatment effect, detecting an effect where none should exist would raise concerns about the model's validity (and hence about its estimate). To formally assess this, we define the following null hypothesis: the estimated true treatment effect $ATE_{\mathcal{G}}$ is not statistically distinguishable from a (random noise) placebo treatment $ATE_{\text{placebo}}$.  

To test this hypothesis we repeat the random reallocation process $n$ times, thus generating $n = 1600$ individual placebo datasets and compute the placebo treatment effect $ATE_{i, \text{placebo}}$ for every $i$-th dataset. These are then used to construct a reference distribution under the null. A p-value is then computed by comparing the magnitude of $ATE_{\mathcal{G}}$ to the empirical distribution of placebo effects (see section \ref{SM_p_val_placebo} in supplementary material). 

Summary statistics for the placebo distribution are presented in the third column of Table \ref{table:refutation}. As shown, both the mean and median of $ATE_{\text{placebo}}$ are close to zero (around $-5$ kWh/year), whereas the estimated true effect $ATE_{\mathcal{G}}$ is substantially larger by comparison. The very low p-value suggests that the observed effect is highly unlikely under the null hypothesis, providing strong evidence to reject it.

\begin{table}[h!]
\caption{\small Robustness checks for the estimated average treatment effect ($ATE_{\mathcal{G}}$). The placebo test (third column) assesses whether spurious effects emerge when treatment values are randomly reassigned, simulating a null scenario. The subsample test (fourth column) evaluates estimate stability across random subsamples ($\approx40\%$ of the full training dataset). Reported are means, medians, 1-99\% confidence intervals, p-values, and sample sizes. The estimated true effect, shown in the final column, is substantially larger than placebo estimates and consistent with the empirical distribution of treatment effect obtained from random subsamples of the training dataset.}
\centering
\small
\label{table:refutation}
\begin{tabular}{lcccc}
\hline
& Unit & $ATE_{\text{placebo}}$  & $ATE_{\text{sub}}$  & $ATE_{\mathcal{G}}$ (baseline) \\
&  &   &  & [kWh/yr] \\
\hline
Mean & [kWh/yr] & -5.2 & -2935.5 & \multirow{5}{*}{-2980.1} \\
Median & [kWh/yr] & -5.5 & -2935.6 & \\
CI (1\%; 99\%) & [kWh/yr]  & (-66.5; 51.6) & (-3004.1; -2872.2) & \\
p-value\textsuperscript & -- & 0.000625 & 0.1274 & \\
Sample size ($n$) & -- & 1600 & 1800 & \\
\hline
\end{tabular}
\end{table}

\subsection{Data subsample test}
\label{section:data_susample_test}
The second refutation strategy assesses the robustness of the estimated average treatment effect to variations in the training data by employing repeated random subsampling. Specifically, we generate $n=1800$ randomly drawn subsamples of the training dataset, each comprising $\approx40\%$ of the units in the full dataset, and compute the corresponding treatment effect estimate, $ATE_{i, \text{sub}}$, for each subsample $i$. This process yields an empirical distribution of subsample-based $ATEs$, which is then compared to the $ATE$ estimated on the full dataset, i.e. $ATE_{\mathcal{G}}$. 

The intuition behind this test is that, if the full-sample estimate is stable and not unduly influenced by any particular segment of the data, it should lie somewhere within the empirical distribution formed by the subsample estimates. In contrast, a large deviation may signal sensitivity to specific data partitions, calling into question the generalisability of the result. 

Formally, the null hypothesis for this test states that: the treatment effect estimated from the full dataset is not significantly different from the $ATEs$ estimated on random subsamples. Thus, in this context, a high p-value is desirable (see section \ref{SM_p_val_subsample} in supplementary material), as it implies that any observed deviation of $ATE_{\mathcal{G}}$ from the distribution of $ATE_{i, \text{sub}}$ can plausibly be attributed to random sampling variation. Retaining the null hypothesis in this case strengthens the credibility of the causal estimate, indicating that it is not overly sensitive to the particular composition of the training data. Summary statistics for this test are reported in the forth column of Table \ref{table:refutation}, where it can be seen how the estimated true effect ($-2980.1$ kWh/yr) is well within the confidence interval of the empirical distribution for $ATE_{\text{sub}}$.

\subsection{Sensitivity analysis to unobserved confounder}
\begin{figure}[ht]
    \centering
    \includegraphics[width=13.0cm]{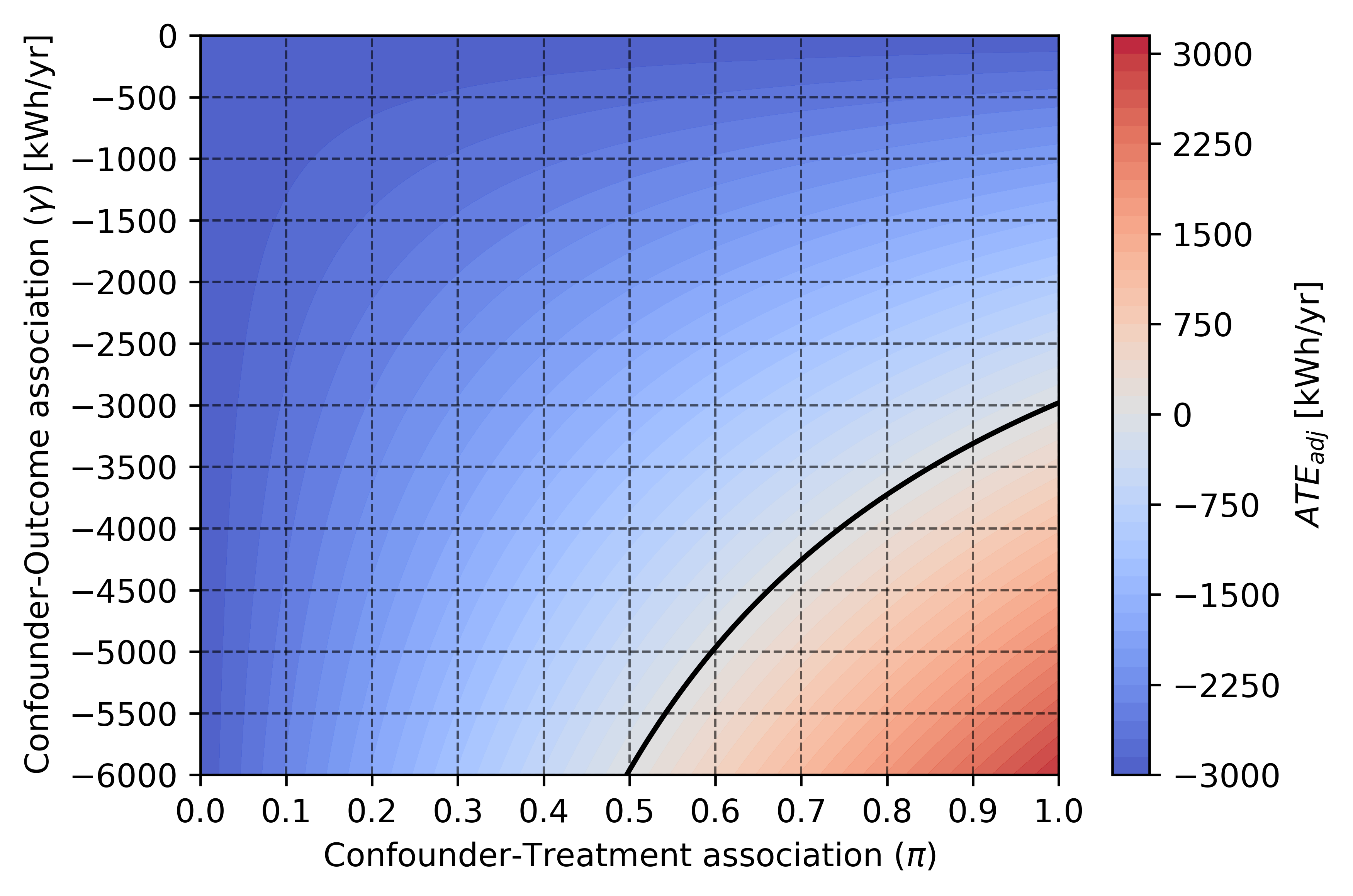}
    \caption{\small Sensitivity analysis of the estimated average treatment effect ($ATE_{\mathcal{G}}$) to a hypothetical unmeasured confounder $U$. The heat-map shows the adjusted treatment effect, $ATE_{\text{adj}} = ATE_{\mathcal{G}} - \pi \cdot \gamma$, where $\pi$ denotes the confounder–treatment association (increase in probability of dwelling insulation attributable to the confounder), and $\gamma$ denotes the confounder–outcome association (change in annual gas consumption attributable to the confounder, independent of insulation). Shading represents the magnitude of the adjusted ATE. The solid black contour represents the tipping line, i.e. the locus of ($\pi, \gamma$) combinations that would exactly explain away the estimated effect of insulation ($ATE_{\mathcal{G}} = -2980.1$ kWh/year).}
    \label{fig:sensitivity}
\end{figure}

While the placebo and subsample tests assess robustness empirically within the observed data, a more fundamental concern is potential unmeasured confounding. By definition, such confounding cannot be directly tested, yet it may still bias causal estimates if an omitted variable influences both treatment assignment and outcomes ($X \leftarrow U\rightarrow Y_0$). To quantify this risk, we conduct a sensitivity analysis, which evaluates how strong such a confounder would have to be in order to fully explain away the estimated treatment effect $ATE_{\mathcal{G}}$. Although the existence of such a confounder is hypothetical and most importantly it's unknowable (it might represent one, several, or none of the underlying factors), we will refer to it as \emph{energy consciousness} of occupants, so to make the following exposition more intuitive. Let say conscious households are more likely to choose residing in insulated dwellings (through selective renting, purchasing, or investing in upgrades) and may also consume less gas for heating, independently of insulation. We formalise this using a standard linear omitted-variable bias (OVB) formula.
Let $U \in \{0,1\}$ denote a binary energy-consciousness indicator, with $U=1$ indicating ``conscious'' households, and suppose $U$ affects the probability of dwelling insulation $X$ by:
\begin{equation}
\label{eq:pi}
    \pi = P(X=true \mid U=1) - P(X=true \mid U=0)
\end{equation}
whereby $\pi$ represents the confounder-treatment association strength (i.e., the effect of energy consciousness on the probability of insulation). Similarly, let $U$ affect the expected annual gas consumption $Y_0$ by:
\begin{equation}
\label{eq:gamma}
    \gamma = \mathbb{E}(Y_0 \mid U=1) - \mathbb{E}(Y_0 \mid U=0)
\end{equation}
where $\gamma$ represents the confounder–outcome association strength (i.e., the effect of energy consciousness on gas consumption, independent of insulation).
If $U$ exists, the estimated $ATE_{\mathcal{G}}$ will be biased by the quantity $\gamma \pi$, so that the adjusted effect is expressed as follows \cite{Wooldridge2013}:
\begin{equation}
\label{eq:biasedATE}
    ATE_{\text{adj}} = ATE_{\mathcal{G}} - \gamma\pi
\end{equation}
Figure \ref{fig:sensitivity} shows $ATE_{\text{adj}}$ across a grid of ($\pi, \gamma$) value pairs. Shading represents the magnitude of the adjusted effect, while the solid contour marks the tipping line where $ATE_{\text{adj}} = 0$. This line identifies all combinations of confounder–treatment and confounder–outcome association strengths sufficient to nullify the estimated effect of insulation ($ATE_{\mathcal{G}} = -2980.1$ kWh/y). 
To illustrate, if energy-conscious households were three times more likely to choose living in insulated dwellings ($\pi = 0.75 - 0.25 = 0.5$) then energy consciousness would also need to reduce consumption independently by $-5960$ kWh/yr ($\approx40\%$ of average baseline demand) to eliminate the insulation effect. For more modest confounder–treatment associations the confounder–outcome association required to explain away\footnote{To `explain away' an estimated effect means to show that the entire observed association between two variables can be attributed to (or is fully accounted for by) a third, unobserved variable (a confounder). In this context, it implies driving the estimated true causal effect to zero.} the estimated effect becomes unrealistically large. For instance, assuming conscious households are only 1/3 more likely to choose insulated dwellings compared to uncaring ones ($\pi = 0.6 - 0.4 = 0.2$) then the tipping line implies that consciousness independently drives these households to save as much as $\gamma\approx -14900$ kWh/year in gas, which exceeds the average annual baseline gas consumption of English households. 
Similarly, if energy consciousness, $U$, reduces consumption by a more modest $\gamma \approx -3000$ kWh/yr, $U$ would need to be an almost deterministic predictor of insulation, i.e., $\pi = P(X=true \mid U=1) - P(X=true \mid U=0) \approx 1$. 
More generally, any realistic value for $\pi$ would require an implausibly large effect of $U$ on $Y_0$, and any realistic effect of $U$ on $Y_0$ would require an implausibly strong association with insulation ($\pi$).  Thus, within plausible ranges for both $\pi$ and $\gamma$, unmeasured confounding cannot fully explain away the estimated effect $ATE_{\mathcal{G}}$. This sensitivity analysis therefore provides further quantitative evidence for the robustness of our causal conclusions.

\section{Conclusions}\label{sec_conclusions}

The study advances understanding of the causal impacts in domestic energy efficiency interventions by combining a novel methodological framework with nationally representative data. Specifically, we employed a (semi-Markovian) causal Bayesian network and do-calculus, to estimate the causal effect of external wall insulation on household gas demand for space heating. Building on earlier studies using econometric methods (such as difference-in-differences and matching techniques) with the National Energy Efficiency Data (NEED), our methodology explicitly formalises causal assumptions through a graphical model. This enables transparent causal effect identification based on a \emph{sufficient} adjustment set and provides a flexible basis for robust estimation of both average and Conditional Treatment Effects.

Crucially, this work also demonstrates the feasibility of using the English Housing Survey - Fuel Poverty (EHS-FP) dataset for causal inference at a national scale. Because the EHS-FP includes household-level income data and provides a nationally representative random sample, our estimates generalise to the entire English housing stock of approximately 22.6 million dwellings. This represents a significant advancement over earlier work, which may be limited by potential self-selection bias from programme-based samples (i.e., NEED datasets) and by the use of area-level deprivation proxies rather than household-level data.

Our findings indicate that external wall insulation reduces household gas consumption by approximately 19\% on average relative to pre-intervention gas consumption levels ($\approx -2980$ kWh/year). However, Conditional Average Treatment Effect (CATE) estimates reveal pronounced heterogeneity depending on pre-existing levels of energy burden (defined as the share of household income spend on energy). Households in the lowest burden strata (1.5–2.5\% of income spent on energy) achieve the largest savings ($\approx -3720$ kWh/year, $\approx -26\%$). In contrast, those in the highest energy burden category ($>$ 11\%) realise only marginal reductions ($\approx -500$ kWh/year, $\approx -3\%$). These divergent outcomes are best understood as behaviourally rational responses to economic constraints: households already underheating prior to the intervention likely reallocate the financial relief from reduced bills toward higher thermal comfort rather than maximising energy savings. Far from being wasteful, such behavioural responses constitute rational adjustments in contexts of fuel poverty, with likely co-benefits for health and well-being.

The contribution of this study is therefore twofold. Substantively, it provides country-wide causal evidence that the impacts of energy efficiency retrofits are strongly conditioned by household-level energy burden, revealing a structural trade-off between carbon mitigation and social welfare gains. Methodologically, it demonstrates the applicability of causal graphical models for energy policy evaluation for the built environment, and that the EHS-FP dataset can serve as a robust basis for causal inference across the 22.6 million dwellings constituting the English housing stock. 

Policy implications follow directly. For households with low energy burden, insulation measures deliver the expected operational carbon reductions in line with engineering model predictions. For high-burden households, however, the principal benefits lie in alleviating thermal deprivation and improving well-being, with only limited net carbon savings. This finding reveals a structural trade-off that challenges the prevailing ``win–win'' framing of energy efficiency as simultaneously maximising climate and social objectives. We argue that policy evaluation frameworks should explicitly integrate social welfare indicators such as thermal comfort, health outcomes, and equity alongside carbon accounting. Recognising and valuing these co-benefits is essential to accurately capture the societal impact of energy efficiency interventions across the full housing stock and to support a more just and sustainable low-carbon transition.

\section{Data and code availability}\label{sec_DC_aval}
All data and code developed for this study are publicly available and can be accessed via the referenced GitHub repository \cite{github_repo}. 

\section{Acknowledgements}
B.D. and L.K. express their gratitude to the School of Computing, Engineering \& the Built Environment (SCEBE) at Edinburgh Napier University for providing the support and resources necessary for the completion of this research. 

\bibliography{sn-bibliography}


\begin{thebibliography}{78}
\ifx \bisbn   \undefined \def \bisbn  #1{ISBN #1}\fi
\ifx \binits  \undefined \def \binits#1{#1}\fi
\ifx \bauthor  \undefined \def \bauthor#1{#1}\fi
\ifx \batitle  \undefined \def \batitle#1{#1}\fi
\ifx \bjtitle  \undefined \def \bjtitle#1{#1}\fi
\ifx \bvolume  \undefined \def \bvolume#1{\textbf{#1}}\fi
\ifx \byear  \undefined \def \byear#1{#1}\fi
\ifx \bissue  \undefined \def \bissue#1{#1}\fi
\ifx \bfpage  \undefined \def \bfpage#1{#1}\fi
\ifx \blpage  \undefined \def \blpage #1{#1}\fi
\ifx \burl  \undefined \def \burl#1{\textsf{#1}}\fi
\ifx \doiurl  \undefined \def \doiurl#1{\url{https://doi.org/#1}}\fi
\ifx \betal  \undefined \def \betal{\textit{et al.}}\fi
\ifx \binstitute  \undefined \def \binstitute#1{#1}\fi
\ifx \binstitutionaled  \undefined \def \binstitutionaled#1{#1}\fi
\ifx \bctitle  \undefined \def \bctitle#1{#1}\fi
\ifx \beditor  \undefined \def \beditor#1{#1}\fi
\ifx \bpublisher  \undefined \def \bpublisher#1{#1}\fi
\ifx \bbtitle  \undefined \def \bbtitle#1{#1}\fi
\ifx \bedition  \undefined \def \bedition#1{#1}\fi
\ifx \bseriesno  \undefined \def \bseriesno#1{#1}\fi
\ifx \blocation  \undefined \def \blocation#1{#1}\fi
\ifx \bsertitle  \undefined \def \bsertitle#1{#1}\fi
\ifx \bsnm \undefined \def \bsnm#1{#1}\fi
\ifx \bsuffix \undefined \def \bsuffix#1{#1}\fi
\ifx \bparticle \undefined \def \bparticle#1{#1}\fi
\ifx \barticle \undefined \def \barticle#1{#1}\fi
\bibcommenthead
\ifx \bconfdate \undefined \def \bconfdate #1{#1}\fi
\ifx \botherref \undefined \def \botherref #1{#1}\fi
\ifx \url \undefined \def \url#1{\textsf{#1}}\fi
\ifx \bchapter \undefined \def \bchapter#1{#1}\fi
\ifx \bbook \undefined \def \bbook#1{#1}\fi
\ifx \bcomment \undefined \def \bcomment#1{#1}\fi
\ifx \oauthor \undefined \def \oauthor#1{#1}\fi
\ifx \citeauthoryear \undefined \def \citeauthoryear#1{#1}\fi
\ifx \endbibitem  \undefined \def \endbibitem {}\fi
\ifx \bconflocation  \undefined \def \bconflocation#1{#1}\fi
\ifx \arxivurl  \undefined \def \arxivurl#1{\textsf{#1}}\fi
\csname PreBibitemsHook\endcsname

\bibitem[\protect\citeauthoryear{Barrett et~al.}{2022}]{barrett2022energy}
\begin{barticle}
\bauthor{\bsnm{Barrett}, \binits{J.}},
\bauthor{\bsnm{Pye}, \binits{S.}},
\bauthor{\bsnm{Betts-Davies}, \binits{S.}},
\bauthor{\bsnm{Broad}, \binits{O.}},
\bauthor{\bsnm{Price}, \binits{J.}},
\bauthor{\bsnm{Eyre}, \binits{N.}},
\bauthor{\bsnm{Anable}, \binits{J.}},
\bauthor{\bsnm{Brand}, \binits{C.}},
\bauthor{\bsnm{Bennett}, \binits{G.}},
\bauthor{\bsnm{Carr-Whitworth}, \binits{R.}}, \betal:
\batitle{Energy demand reduction options for meeting national zero-emission targets in the {United Kingdom}}.
\bjtitle{Nature Energy}
\bvolume{7}(\bissue{8}),
\bfpage{726}--\blpage{735}
(\byear{2022})
\doiurl{10.1038/s41560-022-01057-y}
\end{barticle}
\endbibitem

\bibitem[\protect\citeauthoryear{Xia-Bauer et~al.}{2024}]{xia2024comparative}
\begin{barticle}
\bauthor{\bsnm{Xia-Bauer}, \binits{C.}},
\bauthor{\bsnm{Gokarakonda}, \binits{S.}},
\bauthor{\bsnm{Guo}, \binits{S.}},
\bauthor{\bsnm{Filippidou}, \binits{F.}},
\bauthor{\bsnm{Thomas}, \binits{S.}},
\bauthor{\bsnm{Maheshwari}, \binits{J.R.}},
\bauthor{\bsnm{Vishwanathan}, \binits{S.S.}}:
\batitle{Comparative analysis of residential building decarbonization policies in major economies: insights from the {EU, China, and India}}.
\bjtitle{Energy Efficiency}
\bvolume{17}(\bissue{5}),
\bfpage{46}
(\byear{2024})
\doiurl{10.1007/s12053-024-10225-w}
\end{barticle}
\endbibitem

\bibitem[\protect\citeauthoryear{Broin et~al.}{2015}]{broin2015energy}
\begin{barticle}
\bauthor{\bsnm{Broin}, \binits{E.{\'O}.}},
\bauthor{\bsnm{N{\"a}ss{\'e}n}, \binits{J.}},
\bauthor{\bsnm{Johnsson}, \binits{F.}}:
\batitle{Energy efficiency policies for space heating in {EU} countries: A panel data analysis for the period 1990--2010}.
\bjtitle{Applied Energy}
\bvolume{150},
\bfpage{211}--\blpage{223}
(\byear{2015})
\doiurl{10.1016/j.apenergy.2015.03.063}
\end{barticle}
\endbibitem

\bibitem[\protect\citeauthoryear{Economidou et~al.}{2020}]{economidou2020review}
\begin{barticle}
\bauthor{\bsnm{Economidou}, \binits{M.}},
\bauthor{\bsnm{Todeschi}, \binits{V.}},
\bauthor{\bsnm{Bertoldi}, \binits{P.}},
\bauthor{\bsnm{D'Agostino}, \binits{D.}},
\bauthor{\bsnm{Zangheri}, \binits{P.}},
\bauthor{\bsnm{Castellazzi}, \binits{L.}}:
\batitle{Review of 50 years of {EU} energy efficiency policies for buildings}.
\bjtitle{Energy and buildings}
\bvolume{225},
\bfpage{110322}
(\byear{2020})
\doiurl{10.1016/j.enbuild.2020.110322}
\end{barticle}
\endbibitem

\bibitem[\protect\citeauthoryear{McAndrew et~al.}{2021}]{mcandrew2021household}
\begin{barticle}
\bauthor{\bsnm{McAndrew}, \binits{R.}},
\bauthor{\bsnm{Mulcahy}, \binits{R.}},
\bauthor{\bsnm{Gordon}, \binits{R.}},
\bauthor{\bsnm{Russell-Bennett}, \binits{R.}}:
\batitle{Household energy efficiency interventions: A systematic literature review}.
\bjtitle{Energy Policy}
\bvolume{150},
\bfpage{112136}
(\byear{2021})
\doiurl{10.1016/j.enpol.2021.112136}
\end{barticle}
\endbibitem

\bibitem[\protect\citeauthoryear{{International Energy Agency}}{2022}]{iea2022netzero}
\begin{botherref}
\oauthor{\bsnm{{International Energy Agency}}}:
Net Zero by 2050: A Roadmap for the Global Energy Sector.
IEA Report.
Available at: \url{https://www.iea.org/reports/net-zero-by-2050}
(2022)
\end{botherref}
\endbibitem

\bibitem[\protect\citeauthoryear{on~Climate~Change}{2023}]{ipcc2023synthesis}
\begin{botherref}
\oauthor{\bsnm{Climate~Change}, \binits{I.P.}}:
Climate Change 2023: Synthesis Report.
IPCC Sixth Assessment Report.
Available at: \url{https://www.ipcc.ch/report/ar6/syr/}
(2023)
\end{botherref}
\endbibitem

\bibitem[\protect\citeauthoryear{{HM Government}}{2021}]{uknetzero2050}
\begin{botherref}
\oauthor{\bsnm{{HM Government}}}:
UK Net Zero Strategy: Build Back Greener.
UK Government Policy Paper.
Available at: \url{https://www.gov.uk/government/publications/net-zero-strategy}
(2021)
\end{botherref}
\endbibitem

\bibitem[\protect\citeauthoryear{Ballesteros-Arjona et~al.}{2022}]{ballesteros2022effects}
\begin{barticle}
\bauthor{\bsnm{Ballesteros-Arjona}, \binits{V.}},
\bauthor{\bsnm{Oliveras}, \binits{L.}},
\bauthor{\bsnm{Mu{\~n}oz}, \binits{J.B.}},
\bauthor{\bsnm{Labry~Lima}, \binits{A.O.}},
\bauthor{\bsnm{Carrere}, \binits{J.}},
\bauthor{\bsnm{Ruiz}, \binits{E.M.}},
\bauthor{\bsnm{Peralta}, \binits{A.}},
\bauthor{\bsnm{Le{\'o}n}, \binits{A.C.}},
\bauthor{\bsnm{Rodr{\'\i}guez}, \binits{I.M.}},
\bauthor{\bsnm{Daponte-Codina}, \binits{A.}}, \betal:
\batitle{What are the effects of energy poverty and interventions to ameliorate it on people's health and well-being?: A scoping review with an equity lens}.
\bjtitle{Energy Research \& Social Science}
\bvolume{87},
\bfpage{102456}
(\byear{2022})
\doiurl{10.1016/j.erss.2021.102456}
\end{barticle}
\endbibitem

\bibitem[\protect\citeauthoryear{Mei and Seo}{2024}]{mei2024relationships}
\begin{barticle}
\bauthor{\bsnm{Mei}, \binits{X.}},
\bauthor{\bsnm{Seo}, \binits{B.K.}}:
\batitle{The relationships among housing, energy poverty, and health: A scoping review}.
\bjtitle{Energy for Sustainable Development}
\bvolume{83},
\bfpage{101568}
(\byear{2024})
\doiurl{10.1016/j.esd.2024.101568}
\end{barticle}
\endbibitem

\bibitem[\protect\citeauthoryear{Moorcroft et~al.}{2025}]{moorcroft2025damp}
\begin{barticle}
\bauthor{\bsnm{Moorcroft}, \binits{C.}},
\bauthor{\bsnm{Whitehouse}, \binits{A.}},
\bauthor{\bsnm{Grigg}, \binits{J.}}:
\batitle{Damp and mouldy home: impact on lung health in childhood}.
\bjtitle{Archives of Disease in Childhood}
(\byear{2025})
\doiurl{10.1136/archdischild-2023-326035}
\end{barticle}
\endbibitem

\bibitem[\protect\citeauthoryear{Singh et~al.}{2022}]{singh2022estimating}
\begin{barticle}
\bauthor{\bsnm{Singh}, \binits{A.}},
\bauthor{\bsnm{Mizdrak}, \binits{A.}},
\bauthor{\bsnm{Daniel}, \binits{L.}},
\bauthor{\bsnm{Blakely}, \binits{T.}},
\bauthor{\bsnm{Baker}, \binits{E.}},
\bauthor{\bsnm{Fleitas~Alfonzo}, \binits{L.}},
\bauthor{\bsnm{Bentley}, \binits{R.}}:
\batitle{Estimating cardiovascular health gains from eradicating indoor cold in australia}.
\bjtitle{Environmental health}
\bvolume{21}(\bissue{1}),
\bfpage{54}
(\byear{2022})
\doiurl{10.1186/s12940-022-00865-9}
\end{barticle}
\endbibitem

\bibitem[\protect\citeauthoryear{Harris et~al.}{2010}]{harris2010health}
\begin{botherref}
\oauthor{\bsnm{Harris}, \binits{J.}},
\oauthor{\bsnm{Hall}, \binits{J.}},
\oauthor{\bsnm{Meltzer}, \binits{H.}},
\oauthor{\bsnm{Jenkins}, \binits{R.}},
\oauthor{\bsnm{Oreszczyn}, \binits{T.}},
\oauthor{\bsnm{McManus}, \binits{S.}}:
Health, mental health and housing conditions in England.
{NatCen}.
Available at: \url{https://beatcold.org.uk/wp-content/uploads/2011/08/health-mental-health-and-housing-conditions-in-england.pdf}
(2010)
\end{botherref}
\endbibitem

\bibitem[\protect\citeauthoryear{{Office for National Statistics}}{2021}]{ons2021excess}
\begin{botherref}
\oauthor{\bsnm{{Office for National Statistics}}}:
Excess Winter Mortality in England and Wales: 2020 to 2021 (Provisional) and 2019 to 2020 (Final).
\url{https://www.ons.gov.uk/releases/excesswintermortalityinenglandandwales2020to2021provisionaland2019to2020final}.
Accessed: May 2025
(2021)
\end{botherref}
\endbibitem

\bibitem[\protect\citeauthoryear{Lee et~al.}{2022}]{lee2022fuel}
\begin{botherref}
\oauthor{\bsnm{Lee}, \binits{A.}},
\oauthor{\bsnm{Sinha}, \binits{I.}},
\oauthor{\bsnm{Boyce}, \binits{T.}},
\oauthor{\bsnm{Allen}, \binits{J.}},
\oauthor{\bsnm{Goldblatt}, \binits{P.}}:
{Fuel Poverty, Cold Homes and Health Inequalities}.
{Institute of Health Equity},
London.
Available at: \url{https://www.instituteofhealthequity.org/resources-reports/fuel-poverty-cold-homes-and-health-inequalities-in-the-uk/read-the-report.pdf}
(2022)
\end{botherref}
\endbibitem

\bibitem[\protect\citeauthoryear{Haines et~al.}{2009}]{haines2009public}
\begin{barticle}
\bauthor{\bsnm{Haines}, \binits{A.}},
\bauthor{\bsnm{McMichael}, \binits{A.J.}},
\bauthor{\bsnm{Smith}, \binits{K.R.}},
\bauthor{\bsnm{Roberts}, \binits{I.}},
\bauthor{\bsnm{Woodcock}, \binits{J.}},
\bauthor{\bsnm{Markandya}, \binits{A.}},
\bauthor{\bsnm{Armstrong}, \binits{B.G.}},
\bauthor{\bsnm{Campbell-Lendrum}, \binits{D.}},
\bauthor{\bsnm{Dangour}, \binits{A.D.}},
\bauthor{\bsnm{Davies}, \binits{M.}}, \betal:
\batitle{Public health benefits of strategies to reduce greenhouse-gas emissions: overview and implications for policy makers}.
\bjtitle{The lancet}
\bvolume{374}(\bissue{9707}),
\bfpage{2104}--\blpage{2114}
(\byear{2009})
\doiurl{10.1016/S0140-6736(09)61759-1}
\end{barticle}
\endbibitem

\bibitem[\protect\citeauthoryear{McCoy and Kotsch}{2021}]{mccoy2021quantifying}
\begin{barticle}
\bauthor{\bsnm{McCoy}, \binits{D.}},
\bauthor{\bsnm{Kotsch}, \binits{R.A.}}:
\batitle{Quantifying the distributional impact of energy efficiency measures}.
\bjtitle{The Energy Journal}
\bvolume{42}(\bissue{6}),
\bfpage{121}--\blpage{144}
(\byear{2021})
\doiurl{10.5547/01956574.42.6.dmcc}
\end{barticle}
\endbibitem

\bibitem[\protect\citeauthoryear{Pe{\~n}asco and Anad{\'o}n}{2023}]{penasco2023assessing}
\begin{barticle}
\bauthor{\bsnm{Pe{\~n}asco}, \binits{C.}},
\bauthor{\bsnm{Anad{\'o}n}, \binits{L.D.}}:
\batitle{Assessing the effectiveness of energy efficiency measures in the residential sector gas consumption through dynamic treatment effects: Evidence from england and wales}.
\bjtitle{Energy Economics}
\bvolume{117},
\bfpage{106435}
(\byear{2023})
\doiurl{10.1016/j.eneco.2022.106435}
\end{barticle}
\endbibitem

\bibitem[\protect\citeauthoryear{Aydin et~al.}{2017}]{aydin2017energy}
\begin{barticle}
\bauthor{\bsnm{Aydin}, \binits{E.}},
\bauthor{\bsnm{Kok}, \binits{N.}},
\bauthor{\bsnm{Brounen}, \binits{D.}}:
\batitle{Energy efficiency and household behavior: the rebound effect in the residential sector}.
\bjtitle{{The RAND Journal of Economics}}
\bvolume{48}(\bissue{3}),
\bfpage{749}--\blpage{782}
(\byear{2017})
\doiurl{10.1111/1756-2171.12190}
\end{barticle}
\endbibitem

\bibitem[\protect\citeauthoryear{Cal{\`\i} et~al.}{2016}]{cali2016energy}
\begin{barticle}
\bauthor{\bsnm{Cal{\`\i}}, \binits{D.}},
\bauthor{\bsnm{Osterhage}, \binits{T.}},
\bauthor{\bsnm{Streblow}, \binits{R.}},
\bauthor{\bsnm{M{\"u}ller}, \binits{D.}}:
\batitle{Energy performance gap in refurbished german dwellings: Lesson learned from a field test}.
\bjtitle{Energy and buildings}
\bvolume{127},
\bfpage{1146}--\blpage{1158}
(\byear{2016})
\doiurl{10.1016/j.enbuild.2016.05.020}
\end{barticle}
\endbibitem

\bibitem[\protect\citeauthoryear{Sunikka-Blank and Galvin}{2012}]{sunikkablank2012prebound}
\begin{barticle}
\bauthor{\bsnm{Sunikka-Blank}, \binits{M.}},
\bauthor{\bsnm{Galvin}, \binits{R.}}:
\batitle{Introducing the prebound effect: the gap between performance and actual energy consumption}.
\bjtitle{Building Research \& Information}
\bvolume{40}(\bissue{3}),
\bfpage{260}--\blpage{273}
(\byear{2012})
\doiurl{10.1080/09613218.2012.690952}
\end{barticle}
\endbibitem

\bibitem[\protect\citeauthoryear{Anderson et~al.}{2012}]{anderson2012coping}
\begin{barticle}
\bauthor{\bsnm{Anderson}, \binits{W.}},
\bauthor{\bsnm{White}, \binits{V.}},
\bauthor{\bsnm{Finney}, \binits{A.}}:
\batitle{Coping with low incomes and cold homes}.
\bjtitle{Energy Policy}
\bvolume{49},
\bfpage{40}--\blpage{52}
(\byear{2012})
\doiurl{10.1016/j.enpol.2012.01.002}
\end{barticle}
\endbibitem

\bibitem[\protect\citeauthoryear{San Miguel-Bellod et~al.}{2018}]{san2018relationship}
\begin{barticle}
\bauthor{\bsnm{San~Miguel-Bellod}, \binits{J.}},
\bauthor{\bsnm{Gonz{\'a}lez-Mart{\'\i}nez}, \binits{P.}},
\bauthor{\bsnm{S{\'a}nchez-Ostiz}, \binits{A.}}:
\batitle{The relationship between poverty and indoor temperatures in winter: Determinants of cold homes in social housing contexts from the 40s--80s in {Northern Spain}}.
\bjtitle{Energy and Buildings}
\bvolume{173},
\bfpage{428}--\blpage{442}
(\byear{2018})
\doiurl{10.1016/j.enbuild.2018.05.022}
\end{barticle}
\endbibitem

\bibitem[\protect\citeauthoryear{{ONS}}{2022}]{ons_winter_2022}
\begin{botherref}
\oauthor{\bsnm{{ONS}}}:
{The impact of winter pressures on adults in Great Britain: December 2022}.
{Office for National Statistics}
(2022).
\url{https://www.ons.gov.uk/peoplepopulationandcommunity/wellbeing/articles/theimpactofwinterpressuresonadultsingreatbritain/december2022}
\end{botherref}
\endbibitem

\bibitem[\protect\citeauthoryear{Heckman}{1990}]{heckman1990selection}
\begin{bchapter}
\bauthor{\bsnm{Heckman}, \binits{J.J.}}:
\bctitle{Selection bias and self-selection}.
In: \bbtitle{Econometrics},
pp. \bfpage{201}--\blpage{224}.
\bpublisher{Springer}, \blocation{???}
(\byear{1990}).
\doiurl{10.1007/978-1-349-20570-7_29}
\end{bchapter}
\endbibitem

\bibitem[\protect\citeauthoryear{Pearl}{2009}]{pearl2009causality}
\begin{bbook}
\bauthor{\bsnm{Pearl}, \binits{J.}}:
\bbtitle{Causality -- Models Reasoning and Inference},
\bedition{2nd} edn.
\bpublisher{Cambridge University Press},
\blocation{Cambridge}
(\byear{2009}).
\doiurl{10.1017/CBO9780511803161}
\end{bbook}
\endbibitem

\bibitem[\protect\citeauthoryear{Clelland and Hill}{2019}]{clelland2019deprivation}
\begin{barticle}
\bauthor{\bsnm{Clelland}, \binits{D.}},
\bauthor{\bsnm{Hill}, \binits{C.}}:
\batitle{Deprivation, policy and rurality: the limitations and applications of area-based deprivation indices in scotland}.
\bjtitle{Local Economy}
\bvolume{34}(\bissue{1}),
\bfpage{33}--\blpage{50}
(\byear{2019})
\doiurl{10.1177/026909421982}
\end{barticle}
\endbibitem

\bibitem[\protect\citeauthoryear{Deas et~al.}{2003}]{deas2003measuring}
\begin{barticle}
\bauthor{\bsnm{Deas}, \binits{I.}},
\bauthor{\bsnm{Robson}, \binits{B.}},
\bauthor{\bsnm{Wong}, \binits{C.}},
\bauthor{\bsnm{Bradford}, \binits{M.}}:
\batitle{Measuring neighbourhood deprivation: a critique of the index of multiple deprivation}.
\bjtitle{Environment and Planning C: Government and Policy}
\bvolume{21}(\bissue{6}),
\bfpage{883}--\blpage{903}
(\byear{2003})
\doiurl{10.1068/c0240}
\end{barticle}
\endbibitem

\bibitem[\protect\citeauthoryear{Suppes}{1973}]{suppes1973probabilistic}
\begin{botherref}
\oauthor{\bsnm{Suppes}, \binits{P.}}:
A probabilistic theory of causality.
{British Journal for the Philosophy of Science}
\textbf{24}(4)
(1973)
\end{botherref}
\endbibitem

\bibitem[\protect\citeauthoryear{Greenland et~al.}{1999}]{greenland1999causal}
\begin{barticle}
\bauthor{\bsnm{Greenland}, \binits{S.}},
\bauthor{\bsnm{Pearl}, \binits{J.}},
\bauthor{\bsnm{Robins}, \binits{J.M.}}:
\batitle{Causal diagrams for epidemiologic research}.
\bjtitle{Epidemiology}
\bvolume{10}(\bissue{1}),
\bfpage{37}--\blpage{48}
(\byear{1999})
\end{barticle}
\endbibitem

\bibitem[\protect\citeauthoryear{Griffith et~al.}{2020}]{griffith2020collider}
\begin{barticle}
\bauthor{\bsnm{Griffith}, \binits{G.J.}},
\bauthor{\bsnm{Morris}, \binits{T.T.}},
\bauthor{\bsnm{Tudball}, \binits{M.J.}},
\bauthor{\bsnm{Herbert}, \binits{A.}},
\bauthor{\bsnm{Mancano}, \binits{G.}},
\bauthor{\bsnm{Pike}, \binits{L.}},
\bauthor{\bsnm{Sharp}, \binits{G.C.}},
\bauthor{\bsnm{Sterne}, \binits{J.}},
\bauthor{\bsnm{Palmer}, \binits{T.M.}},
\bauthor{\bsnm{Davey~Smith}, \binits{G.}}, \betal:
\batitle{Collider bias undermines our understanding of covid-19 disease risk and severity}.
\bjtitle{Nature communications}
\bvolume{11}(\bissue{1}),
\bfpage{5749}
(\byear{2020})
\doiurl{10.1038/s41467-020-19478-2}
\end{barticle}
\endbibitem

\bibitem[\protect\citeauthoryear{Tennant et~al.}{2021}]{tennant2021use}
\begin{barticle}
\bauthor{\bsnm{Tennant}, \binits{P.W.}},
\bauthor{\bsnm{Murray}, \binits{E.J.}},
\bauthor{\bsnm{Arnold}, \binits{K.F.}},
\bauthor{\bsnm{Berrie}, \binits{L.}},
\bauthor{\bsnm{Fox}, \binits{M.P.}},
\bauthor{\bsnm{Gadd}, \binits{S.C.}},
\bauthor{\bsnm{Harrison}, \binits{W.J.}},
\bauthor{\bsnm{Keeble}, \binits{C.}},
\bauthor{\bsnm{Ranker}, \binits{L.R.}},
\bauthor{\bsnm{Textor}, \binits{J.}}, \betal:
\batitle{Use of directed acyclic graphs ({DAG}s) to identify confounders in applied health research: review and recommendations}.
\bjtitle{International journal of epidemiology}
\bvolume{50}(\bissue{2}),
\bfpage{620}--\blpage{632}
(\byear{2021})
\doiurl{10.1093/ije/dyaa213}
\end{barticle}
\endbibitem

\bibitem[\protect\citeauthoryear{Spirtes}{2005}]{spirtes2005graphical}
\begin{barticle}
\bauthor{\bsnm{Spirtes}, \binits{P.}}:
\batitle{Graphical models, causal inference, and econometric models}.
\bjtitle{Journal of Economic Methodology}
\bvolume{12}(\bissue{1}),
\bfpage{3}--\blpage{34}
(\byear{2005})
\doiurl{10.1080/1350178042000330887}
\end{barticle}
\endbibitem

\bibitem[\protect\citeauthoryear{Heckman and Pinto}{2024}]{heckman2024econometric}
\begin{barticle}
\bauthor{\bsnm{Heckman}, \binits{J.}},
\bauthor{\bsnm{Pinto}, \binits{R.}}:
\batitle{Econometric causality: The central role of thought experiments}.
\bjtitle{Journal of Econometrics}
\bvolume{243}(\bissue{1-2}),
\bfpage{105719}
(\byear{2024})
\doiurl{10.1016/j.jeconom.2024.105719}
\end{barticle}
\endbibitem

\bibitem[\protect\citeauthoryear{H{\"u}nermund and Bareinboim}{2025}]{hunermund2025causal}
\begin{barticle}
\bauthor{\bsnm{H{\"u}nermund}, \binits{P.}},
\bauthor{\bsnm{Bareinboim}, \binits{E.}}:
\batitle{Causal inference and data fusion in econometrics}.
\bjtitle{The Econometrics Journal}
\bvolume{28}(\bissue{1}),
\bfpage{41}--\blpage{82}
(\byear{2025})
\doiurl{10.1093/ectj/utae008}
\end{barticle}
\endbibitem

\bibitem[\protect\citeauthoryear{Runge et~al.}{2019}]{runge2019inferring}
\begin{barticle}
\bauthor{\bsnm{Runge}, \binits{J.}},
\bauthor{\bsnm{Bathiany}, \binits{S.}},
\bauthor{\bsnm{Bollt}, \binits{E.}},
\bauthor{\bsnm{Camps-Valls}, \binits{G.}},
\bauthor{\bsnm{Coumou}, \binits{D.}},
\bauthor{\bsnm{Deyle}, \binits{E.}},
\bauthor{\bsnm{Glymour}, \binits{C.}},
\bauthor{\bsnm{Kretschmer}, \binits{M.}},
\bauthor{\bsnm{Mahecha}, \binits{M.D.}},
\bauthor{\bsnm{Mu{\~n}oz-Mar{\'\i}}, \binits{J.}}, \betal:
\batitle{Inferring causation from time series in earth system sciences}.
\bjtitle{Nature {C}ommunications}
\bvolume{10}(\bissue{1}),
\bfpage{2553}
(\byear{2019})
\doiurl{10.1038/s41467-019-10105-3}
\end{barticle}
\endbibitem

\bibitem[\protect\citeauthoryear{Richens et~al.}{2020}]{richens2020improving}
\begin{barticle}
\bauthor{\bsnm{Richens}, \binits{J.G.}},
\bauthor{\bsnm{Lee}, \binits{C.M.}},
\bauthor{\bsnm{Johri}, \binits{S.}}:
\batitle{Improving the accuracy of medical diagnosis with causal machine learning}.
\bjtitle{Nature {C}ommunications}
\bvolume{11}(\bissue{1}),
\bfpage{3923}
(\byear{2020})
\doiurl{10.1038/s41467-020-17419-7}
\end{barticle}
\endbibitem

\bibitem[\protect\citeauthoryear{Sanchez et~al.}{2022}]{sanchez2022causal}
\begin{barticle}
\bauthor{\bsnm{Sanchez}, \binits{P.}},
\bauthor{\bsnm{Voisey}, \binits{J.P.}},
\bauthor{\bsnm{Xia}, \binits{T.}},
\bauthor{\bsnm{Watson}, \binits{H.I.}},
\bauthor{\bsnm{O’Neil}, \binits{A.Q.}},
\bauthor{\bsnm{Tsaftaris}, \binits{S.A.}}:
\batitle{Causal machine learning for healthcare and precision medicine}.
\bjtitle{{Royal Society Open Science}}
\bvolume{9}(\bissue{8}),
\bfpage{220638}
(\byear{2022})
\doiurl{10.1098/rsos.220638}
\end{barticle}
\endbibitem

\bibitem[\protect\citeauthoryear{Feuerriegel et~al.}{2024}]{feuerriegel2024causal}
\begin{barticle}
\bauthor{\bsnm{Feuerriegel}, \binits{S.}},
\bauthor{\bsnm{Frauen}, \binits{D.}},
\bauthor{\bsnm{Melnychuk}, \binits{V.}},
\bauthor{\bsnm{Schweisthal}, \binits{J.}},
\bauthor{\bsnm{Hess}, \binits{K.}},
\bauthor{\bsnm{Curth}, \binits{A.}},
\bauthor{\bsnm{Bauer}, \binits{S.}},
\bauthor{\bsnm{Kilbertus}, \binits{N.}},
\bauthor{\bsnm{Kohane}, \binits{I.S.}},
\bauthor{\bsnm{Schaar}, \binits{M.}}:
\batitle{Causal machine learning for predicting treatment outcomes}.
\bjtitle{Nature Medicine}
\bvolume{30}(\bissue{4}),
\bfpage{958}--\blpage{968}
(\byear{2024})
\doiurl{10.1038/s41591-024-02902-1}
\end{barticle}
\endbibitem

\bibitem[\protect\citeauthoryear{Pearl}{2019}]{pearl2019limitations}
\begin{bchapter}
\bauthor{\bsnm{Pearl}, \binits{J.}}:
\bctitle{The limitations of opaque learning machines}.
In: \beditor{\bsnm{Brockman}, \binits{J.}} (ed.)
\bbtitle{Possible Minds: Twenty-Five Ways of Looking at AI}.
\bpublisher{Penguin Press},
\blocation{London}
(\byear{2019})
\end{bchapter}
\endbibitem

\bibitem[\protect\citeauthoryear{Hollmann et~al.}{2025}]{hollmann2025accurate}
\begin{barticle}
\bauthor{\bsnm{Hollmann}, \binits{N.}},
\bauthor{\bsnm{M{\"u}ller}, \binits{S.}},
\bauthor{\bsnm{Purucker}, \binits{L.}},
\bauthor{\bsnm{Krishnakumar}, \binits{A.}},
\bauthor{\bsnm{K{\"o}rfer}, \binits{M.}},
\bauthor{\bsnm{Hoo}, \binits{S.B.}},
\bauthor{\bsnm{Schirrmeister}, \binits{R.T.}},
\bauthor{\bsnm{Hutter}, \binits{F.}}:
\batitle{Accurate predictions on small data with a tabular foundation model}.
\bjtitle{Nature}
\bvolume{637}(\bissue{8045}),
\bfpage{319}--\blpage{326}
(\byear{2025})
\doiurl{10.1038/s41586-024-08328-6}
\end{barticle}
\endbibitem

\bibitem[\protect\citeauthoryear{{Department for Business, Energy \& Industrial Strategy}}{2021}]{EHS_FP_2018}
\begin{botherref}
\oauthor{\bsnm{{Department for Business, Energy \& Industrial Strategy}}}:
English Housing Survey: Fuel Poverty Dataset, 2018.
\doiurl{10.5255/UKDA-SN-8655-2} .
[data collection]
\end{botherref}
\endbibitem

\bibitem[\protect\citeauthoryear{{Department for Energy Security and Net Zero}}{2023}]{UKDA9243}
\begin{botherref}
\oauthor{\bsnm{{Department for Energy Security and Net Zero}}}:
Fuel poverty dataset documentation – end user licence.
Technical report,
{UK Data Archive}
(2023).
\url{https://doc.ukdataservice.ac.uk/doc/9243/mrdoc/pdf/9243_2021_fuel_poverty_dataset_documentation_eul.pdf}
\end{botherref}
\endbibitem

\bibitem[\protect\citeauthoryear{Robins and Greenland}{1992}]{robins1992identifiability}
\begin{barticle}
\bauthor{\bsnm{Robins}, \binits{J.M.}},
\bauthor{\bsnm{Greenland}, \binits{S.}}:
\batitle{Identifiability and exchangeability for direct and indirect effects}.
\bjtitle{Epidemiology}
\bvolume{3}(\bissue{2}),
\bfpage{143}--\blpage{155}
(\byear{1992})
\end{barticle}
\endbibitem

\bibitem[\protect\citeauthoryear{Pearl}{2001}]{pearl2001direct}
\begin{bchapter}
\bauthor{\bsnm{Pearl}, \binits{J.}}:
\bctitle{Direct and indirect effects}.
In: \bbtitle{Proceedings of the Seventeenth Conference on Uncertainty in Artificial Intelligence},
\bconflocation{San {F}ransisco, {C}{A}},
pp. \bfpage{411}--\blpage{420}
(\byear{2001})
\end{bchapter}
\endbibitem

\bibitem[\protect\citeauthoryear{Laubach et~al.}{2021}]{laubach2021biologist}
\begin{barticle}
\bauthor{\bsnm{Laubach}, \binits{Z.M.}},
\bauthor{\bsnm{Murray}, \binits{E.J.}},
\bauthor{\bsnm{Hoke}, \binits{K.L.}},
\bauthor{\bsnm{Safran}, \binits{R.J.}},
\bauthor{\bsnm{Perng}, \binits{W.}}:
\batitle{A biologist's guide to model selection and causal inference}.
\bjtitle{Proceedings of the Royal Society B}
\bvolume{288}(\bissue{1943}),
\bfpage{20202815}
(\byear{2021})
\doiurl{10.1098/rspb.2020.2815}
\end{barticle}
\endbibitem

\bibitem[\protect\citeauthoryear{Poppe et~al.}{2025}]{poppe2025develop}
\begin{barticle}
\bauthor{\bsnm{Poppe}, \binits{L.}},
\bauthor{\bsnm{Steen}, \binits{J.}},
\bauthor{\bsnm{Loh}, \binits{W.W.}},
\bauthor{\bsnm{Crombez}, \binits{G.}},
\bauthor{\bsnm{De~Block}, \binits{F.}},
\bauthor{\bsnm{Jacobs}, \binits{N.}},
\bauthor{\bsnm{Tennant}, \binits{P.W.}},
\bauthor{\bsnm{Cauwenberg}, \binits{J.V.}},
\bauthor{\bsnm{Paepe}, \binits{A.L.D.}}:
\batitle{How to develop causal directed acyclic graphs for observational health research: a scoping review}.
\bjtitle{Health Psychology Review}
\bvolume{19}(\bissue{1}),
\bfpage{45}--\blpage{65}
(\byear{2025})
\doiurl{10.1080/17437199.2024.2402809}
\end{barticle}
\endbibitem

\bibitem[\protect\citeauthoryear{Glymour et~al.}{2019}]{glymour2019review}
\begin{barticle}
\bauthor{\bsnm{Glymour}, \binits{C.}},
\bauthor{\bsnm{Zhang}, \binits{K.}},
\bauthor{\bsnm{Spirtes}, \binits{P.}}:
\batitle{Review of causal discovery methods based on graphical models}.
\bjtitle{Frontiers in genetics}
\bvolume{10},
\bfpage{524}
(\byear{2019})
\doiurl{10.3389/fgene.2019.00524}
\end{barticle}
\endbibitem

\bibitem[\protect\citeauthoryear{Cartwright}{1994}]{cartwright1994nocause}
\begin{bchapter}
\bauthor{\bsnm{Cartwright}, \binits{N.}}:
\bctitle{No causes in, no causes out}.
In: \bbtitle{Nature's Capacities and Their Measurement},
pp. \bfpage{39}--\blpage{90}.
\bpublisher{{Oxford University Press}},
\blocation{Oxford}
(\byear{1994}).
\doiurl{10.1093/0198235070.003.0003}
\end{bchapter}
\endbibitem

\bibitem[\protect\citeauthoryear{Koller and Friedman}{2009}]{koller2009probabilistic}
\begin{bbook}
\bauthor{\bsnm{Koller}, \binits{D.}},
\bauthor{\bsnm{Friedman}, \binits{N.}}:
\bbtitle{Probabilistic Graphical Models: Principles and Techniques}.
\bpublisher{MIT press},
\blocation{Cambridge}
(\byear{2009})
\end{bbook}
\endbibitem

\bibitem[\protect\citeauthoryear{Pearl}{1995}]{pearl1995causal}
\begin{barticle}
\bauthor{\bsnm{Pearl}, \binits{J.}}:
\batitle{Causal diagrams for empirical research}.
\bjtitle{Biometrika}
\bvolume{82}(\bissue{4}),
\bfpage{669}--\blpage{688}
(\byear{1995})
\doiurl{10.1093/biomet/82.4.669}
\end{barticle}
\endbibitem

\bibitem[\protect\citeauthoryear{Pearl et~al.}{2016}]{pearl2016causal}
\begin{bbook}
\bauthor{\bsnm{Pearl}, \binits{J.}},
\bauthor{\bsnm{Glymour}, \binits{M.}},
\bauthor{\bsnm{Jewell}, \binits{N.P.}}:
\bbtitle{Causal Inference in Statistics: A Primer}.
\bpublisher{John Wiley \& Sons},
\blocation{Hoboken, NJ}
(\byear{2016})
\end{bbook}
\endbibitem

\bibitem[\protect\citeauthoryear{Montgomery et~al.}{2018}]{montgomery2018conditioning}
\begin{barticle}
\bauthor{\bsnm{Montgomery}, \binits{J.M.}},
\bauthor{\bsnm{Nyhan}, \binits{B.}},
\bauthor{\bsnm{Torres}, \binits{M.}}:
\batitle{How conditioning on post-treatment variables can ruin your experiment and what to do about it}.
\bjtitle{{American Journal of Political Science}}
\bvolume{62}(\bissue{3}),
\bfpage{760}--\blpage{775}
(\byear{2018})
\doiurl{10.1111/ajps.12357}
\end{barticle}
\endbibitem

\bibitem[\protect\citeauthoryear{Pearl}{2015}]{pearl2015conditioning}
\begin{barticle}
\bauthor{\bsnm{Pearl}, \binits{J.}}:
\batitle{Conditioning on post-treatment variables}.
\bjtitle{{Journal of Causal Inference}}
\bvolume{3}(\bissue{1}),
\bfpage{131}--\blpage{137}
(\byear{2015})
\doiurl{10.1515/jci-2015-0005}
\end{barticle}
\endbibitem

\bibitem[\protect\citeauthoryear{Zhang and Poole}{1994}]{zhang1994simple}
\begin{bchapter}
\bauthor{\bsnm{Zhang}, \binits{N.L.}},
\bauthor{\bsnm{Poole}, \binits{D.}}:
\bctitle{A simple approach to bayesian network computations}.
In: \bbtitle{Proc. of the Tenth Canadian Conference on Artificial Intelligence}
(\byear{1994}).
\burl{https://hdl.handle.net/1783.1/757}
\end{bchapter}
\endbibitem

\bibitem[\protect\citeauthoryear{Henderson and Hart}{2012}]{henderson2012bredem}
\begin{botherref}
\oauthor{\bsnm{Henderson}, \binits{J.}},
\oauthor{\bsnm{Hart}, \binits{J.}}:
{B}{R}{E}{D}{E}{M} 2012 -- a technical description of the {B}{R}{E} {D}omestic {E}nergy {M}odel.
Technical report,
{B}uilding {R}esearch {E}stablishment, UK
(2012).
\url{https://files.bregroup.com/bre-co-uk-file-library-copy/filelibrary/bredem/BREDEM-2012-specification.pdf}
\end{botherref}
\endbibitem

\bibitem[\protect\citeauthoryear{{B}{R}{E}}{2017}]{FulePoverty_methodology}
\begin{botherref}
\oauthor{\bsnm{{B}{R}{E}}}:
{F}uel {P}overty -- methodology handbook.
Technical report,
{D}epartment for {B}usiness, {E}nergy and {I}ndustrial {S}trategy, UK
(2017).
\url{https://doc.ukdataservice.ac.uk/doc/8289/mrdoc/pdf/8289_fuel_poverty_methodology_handbook.pdf}
\end{botherref}
\endbibitem

\bibitem[\protect\citeauthoryear{Heckman}{1979}]{heckman1979sample}
\begin{botherref}
\oauthor{\bsnm{Heckman}, \binits{J.J.}}:
Sample selection bias as a specification error.
Econometrica: Journal of the {E}conometric {S}ociety,
153--161
(1979)
\doiurl{https://www.jstor.org/stable/1912352}
\end{botherref}
\endbibitem

\bibitem[\protect\citeauthoryear{Lee}{2005}]{lee2005training}
\begin{botherref}
\oauthor{\bsnm{Lee}, \binits{D.S.}}:
Training, wages, and sample selection: Estimating sharp bounds on treatment effects.
National Bureau of Economic Research Cambridge, Mass., USA
(2005).
\url{https://ssrn.com/abstract=837164}
\end{botherref}
\endbibitem

\bibitem[\protect\citeauthoryear{Bareinboim and Tian}{2015}]{bareinboim2015recovering}
\begin{bchapter}
\bauthor{\bsnm{Bareinboim}, \binits{E.}},
\bauthor{\bsnm{Tian}, \binits{J.}}:
\bctitle{Recovering causal effects from selection bias}.
In: \bbtitle{Proceedings of the AAAI Conference on Artificial Intelligence},
vol. \bseriesno{29}
(\byear{2015}).
\doiurl{10.1609/aaai.v29i1.9679}
\end{bchapter}
\endbibitem

\bibitem[\protect\citeauthoryear{Bareinboim and Pearl}{2016}]{bareinboim2016causal}
\begin{barticle}
\bauthor{\bsnm{Bareinboim}, \binits{E.}},
\bauthor{\bsnm{Pearl}, \binits{J.}}:
\batitle{Causal inference and the data-fusion problem}.
\bjtitle{Proceedings of the National Academy of Sciences}
\bvolume{113}(\bissue{27}),
\bfpage{7345}--\blpage{7352}
(\byear{2016})
\doiurl{10.1073/pnas.151050711}
\end{barticle}
\endbibitem

\bibitem[\protect\citeauthoryear{Bareinboim et~al.}{2022}]{bareinboim2022recovering}
\begin{bchapter}
\bauthor{\bsnm{Bareinboim}, \binits{E.}},
\bauthor{\bsnm{Tian}, \binits{J.}},
\bauthor{\bsnm{Pearl}, \binits{J.}}:
\bctitle{Recovering from selection bias in causal and statistical inference}.
In: \bbtitle{Probabilistic and Causal Inference: The Works of Judea Pearl},
pp. \bfpage{433}--\blpage{450}
(\byear{2022}).
\doiurl{10.1145/3501714.3501740}
\end{bchapter}
\endbibitem

\bibitem[\protect\citeauthoryear{Geiger et~al.}{1990}]{geiger1990d}
\begin{bchapter}
\bauthor{\bsnm{Geiger}, \binits{D.}},
\bauthor{\bsnm{Verma}, \binits{T.}},
\bauthor{\bsnm{Pearl}, \binits{J.}}:
\bctitle{d-separation: From theorems to algorithms}.
In: \bbtitle{Machine Intelligence and Pattern Recognition}
vol. \bseriesno{10},
pp. \bfpage{139}--\blpage{148}.
\bpublisher{Elsevier},
\blocation{Amsterdam}
(\byear{1990}).
\doiurl{10.1016/B978-0-444-88738-2.50018-X}
\end{bchapter}
\endbibitem

\bibitem[\protect\citeauthoryear{Rosenow et~al.}{2022}]{rosenow2022heating}
\begin{barticle}
\bauthor{\bsnm{Rosenow}, \binits{J.}},
\bauthor{\bsnm{Gibb}, \binits{D.}},
\bauthor{\bsnm{Nowak}, \binits{T.}},
\bauthor{\bsnm{Lowes}, \binits{R.}}:
\batitle{Heating up the global heat pump market}.
\bjtitle{Nature Energy}
\bvolume{7}(\bissue{10}),
\bfpage{901}--\blpage{904}
(\byear{2022})
\doiurl{https://www.nature.com/articles/s41560-022-01104-8}
\end{barticle}
\endbibitem

\bibitem[\protect\citeauthoryear{Adan and Fuerst}{2016}]{adan2016energy}
\begin{barticle}
\bauthor{\bsnm{Adan}, \binits{H.}},
\bauthor{\bsnm{Fuerst}, \binits{F.}}:
\batitle{Do energy efficiency measures really reduce household energy consumption? a difference-in-difference analysis}.
\bjtitle{{Energy Efficiency}}
\bvolume{9},
\bfpage{1207}--\blpage{1219}
(\byear{2016})
\doiurl{10.1007/s12053-015-9418-3}
\end{barticle}
\endbibitem

\bibitem[\protect\citeauthoryear{{Department for Business, Energy \& Industrial Strategy (BEIS)}}{2021a}]{beis2021net}
\begin{botherref}
\oauthor{\bsnm{{Department for Business, Energy \& Industrial Strategy (BEIS)}}}:
{Net Zero Strategy: Build Back Greener}.
{HM Government}.
Accessed: 2025-06-20
(2021).
\url{https://assets.publishing.service.gov.uk/media/6194dfa4d3bf7f0555071b1b/net-zero-strategy-beis.pdf}
\end{botherref}
\endbibitem

\bibitem[\protect\citeauthoryear{{Department for Business, Energy \& Industrial Strategy (BEIS)}}{2021b}]{hbs_2021}
\begin{botherref}
\oauthor{\bsnm{{Department for Business, Energy \& Industrial Strategy (BEIS)}}}:
{Heat and Buildings Strategy}.
{HM Government}.
Accessed: 2025-06-20
(2021).
\url{https://assets.publishing.service.gov.uk/media/61d450eb8fa8f54c14eb14e4/6.7408_BEIS_Clean_Heat_Heat___Buildings_Strategy_Stage_2_v5_WEB.pdf}
\end{botherref}
\endbibitem

\bibitem[\protect\citeauthoryear{{Ofgem}}{2025}]{ofgem_eco}
\begin{botherref}
\oauthor{\bsnm{{Ofgem}}}:
{Energy Company Obligation (ECO)}.
\url{https://www.ofgem.gov.uk/environmental-and-social-schemes/energy-company-obligation-eco}.
Accessed: 2025-06-19
(2025)
\end{botherref}
\endbibitem

\bibitem[\protect\citeauthoryear{Nayak et~al.}{2023}]{nayak2023review}
\begin{barticle}
\bauthor{\bsnm{Nayak}, \binits{B.K.}},
\bauthor{\bsnm{Sansaniwal}, \binits{S.K.}},
\bauthor{\bsnm{Mathur}, \binits{J.}},
\bauthor{\bsnm{Chandra}, \binits{T.}},
\bauthor{\bsnm{Garg}, \binits{V.}},
\bauthor{\bsnm{Gupta}, \binits{R.}}:
\batitle{A review of residential building archetypes and their applications to study building energy consumption}.
\bjtitle{{Architectural Science Review}}
\bvolume{66}(\bissue{3}),
\bfpage{187}--\blpage{200}
(\byear{2023})
\doiurl{10.1080/00038628.2023.2193167}
\end{barticle}
\endbibitem

\bibitem[\protect\citeauthoryear{Shen and Wang}{2024}]{shen2024archetype}
\begin{barticle}
\bauthor{\bsnm{Shen}, \binits{P.}},
\bauthor{\bsnm{Wang}, \binits{H.}}:
\batitle{Archetype building energy modeling approaches and applications: A review}.
\bjtitle{Renewable and Sustainable Energy Reviews}
\bvolume{199},
\bfpage{114478}
(\byear{2024})
\doiurl{10.1016/j.rser.2024.114478}
\end{barticle}
\endbibitem

\bibitem[\protect\citeauthoryear{{Department for Energy Security and Net Zero}}{2024}]{DESNZ2023Subnational}
\begin{botherref}
\oauthor{\bsnm{{Department for Energy Security and Net Zero}}}:
Subnational electricity and gas consumption summary report.
Statistical report,
{UK Government}
(December 2024).
Accessed: 2025-05-08.
\url{https://assets.publishing.service.gov.uk/media/6763dd7ebe7b2c675de30820/Subnational-electricity-and-gas-consumption-summary-report-2023.pdf}
\end{botherref}
\endbibitem

\bibitem[\protect\citeauthoryear{{BBC News}}{2023}]{BBC2023}
\begin{botherref}
\oauthor{\bsnm{{BBC News}}}:
Ministers Scrap Green Heating Plans for New Homes.
{BBC}. Accessed: 2025-05-08.
\url{https://www.bbc.co.uk/news/articles/c3e40pe9185o}
\end{botherref}
\endbibitem

\bibitem[\protect\citeauthoryear{Sharma et~al.}{2021}]{sharma2021dowhy}
\begin{barticle}
\bauthor{\bsnm{Sharma}, \binits{A.}},
\bauthor{\bsnm{Syrgkanis}, \binits{V.}},
\bauthor{\bsnm{Zhang}, \binits{C.}},
\bauthor{\bsnm{K{\i}c{\i}man}, \binits{E.}}:
\batitle{Dowhy: Addressing challenges in expressing and validating causal assumptions}.
\bjtitle{arXiv preprint arXiv:2108.13518}
(\byear{2021})
\doiurl{10.48550/arXiv.2108.13518}
\end{barticle}
\endbibitem

\bibitem[\protect\citeauthoryear{Popper}{1934}]{popper1934logic}
\begin{bbook}
\bauthor{\bsnm{Popper}, \binits{K.R.}}:
\bbtitle{The Logic of Scientific Discovery}.
\bpublisher{Routledge},
\blocation{London}
(\byear{1934}).
\doiurl{10.4324/9780203994627}
\end{bbook}
\endbibitem

\bibitem[\protect\citeauthoryear{Eggers et~al.}{2024}]{eggers2024placebo}
\begin{barticle}
\bauthor{\bsnm{Eggers}, \binits{A.C.}},
\bauthor{\bsnm{Tu{\~n}{\'o}n}, \binits{G.}},
\bauthor{\bsnm{Dafoe}, \binits{A.}}:
\batitle{Placebo tests for causal inference}.
\bjtitle{{American Journal of Political Science}}
\bvolume{68}(\bissue{3}),
\bfpage{1106}--\blpage{1121}
(\byear{2024})
\doiurl{10.1111/ajps.12818}
\end{barticle}
\endbibitem

\bibitem[\protect\citeauthoryear{Wooldridge}{2013}]{Wooldridge2013}
\begin{bbook}
\bauthor{\bsnm{Wooldridge}, \binits{J.M.}}:
\bbtitle{Introductory Econometrics: A Modern Approach},
\bedition{5}th edn.
\bpublisher{South-Western Cengage Learning},
\blocation{Boston, MA}
(\byear{2013})
\end{bbook}
\endbibitem

\bibitem[\protect\citeauthoryear{D'Amico}{2025}]{github_repo}
\begin{botherref}
\oauthor{\bsnm{D'Amico}, \binits{B.}}:
Github repository.
Accessed: 2025-04-02
(2025).
\url{https://github.com/bernardinodamico/Households_energy_wellbeing}
\end{botherref}
\endbibitem

\bibitem[\protect\citeauthoryear{{Department for Energy Security and Net Zero}}{2025}]{Gas_billsData}
\begin{botherref}
\oauthor{\bsnm{{Department for Energy Security and Net Zero}}}:
Average annual domestic gas bills by payment type
(2025).
\url{https://assets.publishing.service.gov.uk/media/67e3ed4155239fa04d411fc5/table_231.xlsx}
\end{botherref}
\endbibitem

\end{thebibliography}

\newpage

\clearpage
\renewcommand{\theequation}{S\arabic{equation}}
\renewcommand{\thetable}{S\arabic{table}}
\renewcommand{\thefigure}{S\arabic{figure}}
\renewcommand{\thesection}{S\arabic{section}}
\setcounter{equation}{0}
\setcounter{table}{0}
\setcounter{figure}{0}
\setcounter{section}{0}

\setcounter{page}{0}
\pagenumbering{arabic}

\section*{Supplementary material}\label{sec_SM}

\subsection{Do-calculus}\label{do_calc_rules}
The \emph{calculus of interventions} (or do-calculus) introduced by Pearl \cite{pearl2009causality}, provides a formal framework for reasoning about causal effects in graphical models. The framework enables the derivation of interventional distributions from observational data, under the causal assumptions encoded in the directed acyclic graph $\mathcal{G}$. Central to the framework is the do-operator, $do(X=x)$, which represents an external intervention that sets the variable $X$ to a specific assignment value $x$, effectively removing its dependence on prior causes. Within this setting, do-calculus provides a set of three axiomatic rules for transforming expressions involving post-intervention probability distributions into ordinary conditional probabilities, when permitted by the structure of the graph. The three rules of do-calculus are as follows:

Let $\mathcal{G}$ be a causal graph, and let $X$, $Y$, $Z$ and $W$ be four singleton (or disjoint sets of) variables, then the three rules of do-calculus are as follows:

\begin{itemize}
    \item[] Rule 1, insertion/deletion of observations:
        \begin{equation}
            P(y \mid do(x),z,w)=P(y \mid do(x),w) \mbox{ \ if: \ } \left( Y \indep Z \mid X,W\right)_{\mathcal{G_{\overline{X}}}}
        \end{equation}
    \item[] Rule 2, action/observation exchange: 
        \begin{equation}
            \label{do_calc_rule_2}
            P(y \mid do(x),do(z),w)=P(y \mid do(x),z,w) \mbox{ \ if: \ } \left( Y \indep Z \mid X,W\right)_{\mathcal{G_{\overline{X},\underline{Z}}}}
        \end{equation}
    \item[] Rule 3, insertion/deletion of actions:
        \begin{equation}
            \label{do_calc_rule_3}
            P(y \mid do(x),do(z),w)=P(y \mid do(x),w) \mbox{ \ if: \ } \left( Y \indep Z \mid X,W\right)_{\mathcal{G_{\overline{X},\overline{Z(W)}}}} 
        \end{equation}
\end{itemize}
In words: 
\begin{itemize}
    \item[-] Rule 1 allows the removal of $Z$ from the conditioning set if $Y$ and $Z$ are conditionally independent given $X$ (and optionally, additional variables $W$) in the subgraph obtained by deleting all incoming edges to $X$.
    \item[-] Rule 2 allows replacing an intervention $do(Z)$ with the observation of $Z$, provided that $Y$ and $Z$ are conditionally independent given $X$ (and possibly $W$) in the subgraph where incoming edges to $X$ and outgoing edges from $Z$ are removed.
    \item[-] Rule 3 allows an intervention $do(Z)$ to be removed from the probability expression insofar $Y$ and $Z$ are conditionally independent given $X$ (and possibly $W$) in the subgraph where incoming edges to $X$ and to the variable subset $Z(W)$ are removed. Here, $Z(W)$ denotes the subset of $Z$-nodes that are not ancestors of any $W$-node in $\mathcal{G_{\overline{X}}}$.
\end{itemize}
The rules presented above are general in nature. For clarity and to facilitate the reading of the causal estimator proof in section \ref{sec_covar_specific_eff}, we derive simplified forms of these rules, tailored for the specific model under consideration. In particular, we apply symbolic simplifications to rule 2 and 3 limitedly to interventions involving the individual variable $Z$. When checking for exchangeability of an intervention $do(Z)$ with its observation $Z$ in an expression involving no other interventions, rule 2 (Eq. (\ref{do_calc_rule_2})) reduces to:

\begin{equation}
    \label{do_calc_rule_2_simplified}
    P(y \mid do(z),w)=P(y \mid z,w) \mbox{ \ if: \ } \left( Y \indep Z \mid X,W\right)_{\mathcal{G_{\underline{Z}}}}
\end{equation}
Similarly, when applying rule 3 to an expression involving only the intervention $do(Z)$, and no observed covariates are being conditioned on, then Eq. (\ref{do_calc_rule_3}) simplifies to:

\begin{equation}
    \label{do_calc_rule_3_simplified}
    P(y \mid do(z))=P(y) \mbox{ \ if: \ } \left( Y \indep Z\right)_{\mathcal{G_{\overline{Z}}}}
\end{equation}
These simplifications follow from substituting variable sets with empty sets. In particular, for the original formulation of rule 3 (Eq. (\ref{do_calc_rule_3})) we have that $X=\emptyset$ and $W=\emptyset$, hence the relevant subgraph to check for independency condition is reduced as follows: 
 
\begin{equation}
    \label{empy_set_equivalence}
    \mathcal{G_{\overline{X},\overline{Z(W)}}}=\mathcal{G_{\overline{\emptyset},\overline{Z(\emptyset)}}} =
    \mathcal{G_{\overline{Z(\emptyset)}}}
\end{equation}
Here, $Z(\emptyset)$ denotes the subset of $Z$-nodes that are not ancestors of any variable in the set $W=\emptyset$ relatively to $\mathcal{G_{\overline{\emptyset}}}$ (i.e. to $\mathcal{G}$ itself). Clearly, since no variable can be an ancestor of an empty variable set, all elements of $Z$ trivially satisfy this condition, thus implying that  $\mathcal{G_{\overline{Z(\emptyset)}}} = \mathcal{G_{\overline{Z}}}$ in Eq. (\ref{empy_set_equivalence}), which yields to the simplified form of do-calculus rule 3 stated in  Eq. (\ref{do_calc_rule_3_simplified}).

\subsection{Derivation of estimator formula for covariate-specific causal effects}\label{sec_covar_specific_eff}

\begin{figure}[ht]
    \centering
    \includegraphics[width=14.5cm]{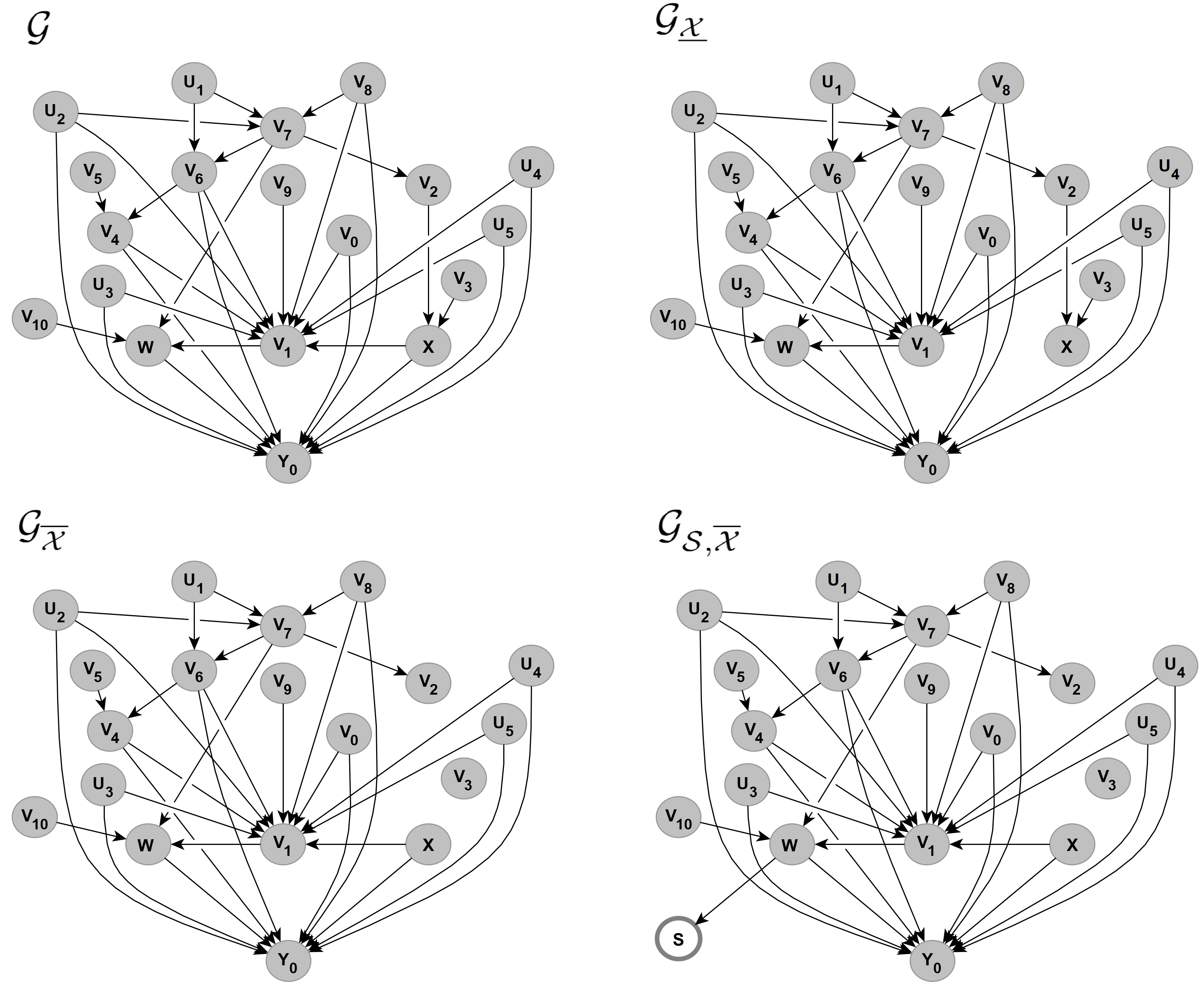}
    \caption{\small The directed acyclic graph $\mathcal{G}$ of the implemented semi-Markovian causal model is shown at the top-left. Subgraphs resulting from selectively removing edges from $\mathcal{G}$ are shown at the top-right ($\mathcal{G_{\underline{X}}}$) and bottom-left ($\mathcal{G_{\overline{X}}}$). Specifically: $\mathcal{G_{\underline{X}}}$ is obtained by removing from $\mathcal{G}$ all directed edges originating from node $X$, whereas $\mathcal{G_{\overline{X}}}$ is obtained by removing all edges in $\mathcal{G}$ pointing to node $X$. The figure on the bottom-right side shows the augmented graph $\mathcal{G_{S}}$ after removing all edges pointing to $X$.}
    \label{fig:subgraphs}
\end{figure}

\begin{proof}[Proof of Eq. (\ref{Eq: covariate_y}).]

We formally define $P(y_0 \mid do(x), w)$ as the post-intervention probability of $Y_0=y_0$ given that an external intervention setting $X=x$ was applied, and that (under such condition) we observe $W=w$. By the law of total probability, this quantity can be expressed as the product (joining) of two \emph{factors} followed by marginalising (summing out) over the variable $V_7$:

\begin{equation}
    \label{Eq: SM_a}
    P(y_0 \mid do(x), w) = \sum_{V_7} P(y_0, v_7 \mid do(x), w) = \sum_{V_7} P(y_0 \mid do(x), w, v_7) P(v_7 \mid do(x), w)
\end{equation}

The first factor on the r.h.s. of Eq. (\ref{Eq: SM_a}) is reduced to an equivalent observational probability distribution by applying rule 2 of Pearl's do-calculus (action/observation exchange) that is Eq. (\ref{do_calc_rule_2_simplified}) here. We are therefore allowed to replace the intervention $do(X=x)$ with the observation $X=x$ insofar the following conditional independency condition is met: 

\begin{equation}
    \label{Eq: SM_b}
    P(y_0 \mid do(x), w, v_7) = P(y_0 \mid x, w, v_7) \mbox{ \ if: \ } \left(Y_0 \indep X \mid W, V_7 \right)_{\mathcal{G_{\underline{X}}}}
\end{equation}

This condition can be verified graphically using the d-separation criterion \cite{geiger1990d}. Specifically, in the subgraph $\mathcal{G_{\underline{X}}}$ (see Figure \ref{fig:subgraphs}), a total of 67 simple\footnote{A simple path is one in which no node is visited more than once.} backdoor paths from $Y_0$ to $X$ were identified via the d-separation algorithm. All of these paths start at $Y_0$ and necessarily pass through the three-node segment $Y_0 ... \leftrightarrows V_7 \rightarrow V_2 \rightarrow X$. This structure is easily verifiable by visual (human) inspection of $\mathcal{G_{\underline{X}}}$.
For any node that immediately precedes $V_7$ in this segment, the connection to $V_7$ forms either a \emph{chain} (e.g. $U_1 \rightarrow V_7 \rightarrow V_2$); or a \emph{fork} (e.g. $V_6 \leftarrow V_7 \rightarrow V_2$). According to the d-separation criterion, conditioning on $V_7$ is sufficient to block all such backdoor paths, as conditioning on the middle node of a chain or a fork structure prevents the flow of information.
Importantly, any collider structure involving $V_7$ as the middle node (e.g. $U_1 \rightarrow V_7 \leftarrow V_8$) are not part of any backdoor path from $Y_0$ to $X$. Consequently, conditioning on $V_7$ does not risk unblocking a collider path, which would otherwise introduce statistical dependency. Here, by unblocking a path, we refer to enabling the propagation of mutual information (e.g. correlation) along that path. Since our objective is to estimate the causal effect of $X$ on $Y_0$, we must ensure that only causal pathways ---such as $X \rightarrow V_1 \rightarrow W \rightarrow Y_0$ and $X \rightarrow Y_0$ in $\mathcal{G}$--- contribute to the observed association. Allowing non-causal associations to propagate through backdoor paths would introduce bias into the effect estimation. By conditioning on $V_7$, we effectively block all such `spurious' associations, ensuring an unbiased causal effect estimate.

Having verified the condition in Eq (\ref{Eq: SM_b}) we now express the second factor on the r.h.s. of Eq. (\ref{Eq: SM_a}) as the ratio:

\begin{equation}
    \label{Eq: SM_c}
    P(v_7 \mid do(x), w) = \frac{P(v_7,w \mid do(x))}{P(w \mid do(x))}
\end{equation}

where the term at the numerator can be factorised as follows:

\begin{equation}
    \label{Eq: SM_d}
    P(v_7, w \mid do(x)) = \sum_{V_2}  P(v_7, w, v_2 \mid do(x)) = \sum_{V_2}  P(v_7, w \mid do(x), v_2) P(v_2 \mid do(x))
\end{equation}

The first factor on the r.h.s. of Eq. (\ref{Eq: SM_d}) is reduced to an observational probability distribution using rule 2 of do-calculus (Eq. (\ref{do_calc_rule_2_simplified}) here):

\begin{equation}
    \label{Eq: SM_e}
    P(v_7, w \mid do(x), v_2) = P(v_7, w \mid x, v_2 ) \mbox{ \ if: \ } \left(V_7, W \indep X \mid V_2 \right)_{\mathcal{G_{\underline{X}}}}
\end{equation}

With reference to the subgraph $\mathcal{G_{\underline{X}}}$ shown in Figure \ref{fig:subgraphs}, by conditioning on $V_2$ we block the only existing backdoor path (a chain) connecting $V_7$ to $X$ ($V_7 \rightarrow V_2 \rightarrow X$) therefore rendering $V_7$ independent of $X$ given $V_2$. To check whether $W$ is also independent of $X$ conditional on $V_2$, we inspect $\mathcal{G_{\underline{X}}}$ for backdoor paths between $W$ and $X$ thus identifying 101 of them. All these paths are reaching $X$ via the three-node segment $W... \leftrightarrows V_7 \rightarrow V_2 \rightarrow X$. As such, being $V_2$ the middle node of this chain segment, conditioning on it blocks all of the 101 backdoor paths $W...\leftrightarrows...X$ therefore rendering $X$ and $W$ conditionally independent given $V_2$.

Having verified the independency condition Eq. (\ref{Eq: SM_e}) we reduce now the second factor in the r.h.s of Eq. (\ref{Eq: SM_d}). To do so, we apply do-calculus rule 3 (insertion/deletion of actions), that is Eq. (\ref{do_calc_rule_3_simplified}) in here:

\begin{equation}
    \label{Eq: SM_f}
    P(v_2 \mid do(x)) = P(v_2) \mbox{ \ if: \ } \left(V_2 \indep X \right)_{\mathcal{G_{\overline{X}}}}
\end{equation}

To determine whether $V_2$ is marginally independent of $X$ in the mutilated graph $\mathcal{G_{\overline{X}}}$ (shown in Figure \ref{fig:subgraphs}) we first identify all existing backdoor paths $X...\leftrightarrows...V_2$, counting a total of 134 such paths. Since we aim to establish marginal (absolute) independence between $X$ and $V_2$, a sufficient condition for this to hold is that every identified backdoor path includes a collider structure along its way where the collider node is not being conditioned on.

As illustrated in the mutilated graph $\mathcal{G_{\overline{X}}}$ shown in Figure \ref{fig:subgraphs}, any backdoor path from $X$ to $V_2$ must first pass through either $V_1$ or $Y_0$ immediately after leaving $X$. Since $Y_0$ has no outgoing edges, any path passing though it is blocked by an unconditioned collider at $Y_0$ (i.e. $X \rightarrow Y_0 \leftarrow ... V_2$). For the remaining backdoor paths passing through $V_1$ they are also blocked due to the unconditioned collider at $V_1$ (i.e. $X \rightarrow V_1 \leftarrow ... V_2$) except for the following path:  

\begin{itemize}
    \item[] $X \rightarrow V_1 \rightarrow W \leftrightarrows... V_2$
\end{itemize}
However, any continuation of this path must next pass through either $V_7$ (via $V_1 \rightarrow W \leftarrow V_7$) or through $Y_0$ (via $V_1 \rightarrow W \rightarrow Y_0 \leftarrow ...$). In both cases,  the presence of an unconditioned collider blocks the path. Thus, no unblocked backdoor paths exist between $X$ and $V_2$, implying that $V_2$ is marginally independent of $X$ in $\mathcal{G_{\overline{X}}}$.

Having reduced both post-intervention factor distributions in the r.h.s. of Eq. (\ref{Eq: SM_d}), we can now express its l.h.s. entirely in terms of observational probability distributions:

\begin{equation}
    \label{Eq: SM_g}
    P(v_7, w \mid do(x)) = \sum_{V_2} P(v_7, w \mid x, v_2 ) P(v_2)
\end{equation}

which gives us a ``do-free'' expression for the numerator term in Eq. (\ref{Eq: SM_c}), hence we now attempt to reduce the factor at the denominator in the same equation. To do so we first express such factor as the product of two more factors followed by marginalising over $V_2$:

\begin{equation}
    \label{Eq: SM_h}
    P(w \mid do(x)) = \sum_{V_2} P(w, v_2 \mid do(x)) = \sum_{V_2} P(w \mid do(x), v_2) P(v_2 \mid do(x))
\end{equation}

nothing that the second term on the r.h.s. of the above Eq. (\ref{Eq: SM_h}) has already been reduced to an observational quantity via Eq. (\ref{Eq: SM_f}), whereas the first r.h.s. term is reduced as follows using rule 2 of do-calculus:

\begin{equation}
    \label{Eq: SM_i}
    P(w \mid do(x), v_2) = P(w \mid x, v_2) \mbox{ \ if: \ } \left(W \indep X \mid V_2 \right)_{\mathcal{G_{\underline{X}}}}
\end{equation}

which holds true insofar we can verify that $W$ is independent of $X$ conditional on $V_2$ in the $\mathcal{G_{\underline{X}}}$ subgraph. This independency condition has already been proved to hold when checking for Eq. (\ref{Eq: SM_e}), hence we can now substitute Eqs. (\ref{Eq: SM_i}) and (\ref{Eq: SM_f}) into Eq. (\ref{Eq: SM_h}):

\begin{equation}
    \label{Eq: SM_l}
    P(w \mid do(x)) = \sum_{V_2} P(w \mid x, v_2) P(v_2)
\end{equation}

which, in turn, is inserted at the denominator of Eq. (\ref{Eq: SM_c}), along with Eq. (\ref{Eq: SM_g}) replacing the numerator instead:

\begin{equation}
    \label{Eq: SM_m}
    P(v_7 \mid do(x), w) = \frac{\sum_{V_2} P(v_7, w \mid x, v_2 ) P(v_2)}{\sum_{V_2} P(w \mid x, v_2) P(v_2)}
\end{equation}

Finally, by inserting the above Eq. (\ref{Eq: SM_m}) and the previously found Eq. (\ref{Eq: SM_b}) into Eq. (\ref{Eq: SM_a}) yields to the estimator formula expressed in Eq. (\ref{Eq: covariate_y}). 

\end{proof}

\newpage

\subsection{Variables description}\label{subsec_variables_descr}

\begin{center}
{\footnotesize
\begin{longtable}{|p{2.4cm}|p{1cm}|c|l|p{4.4cm}|}

\caption{\small Set of observed variables used to construct the graphical causal model. Observed variables, denoted by $V$ (as well as $X, Y_0$ and $W$) are retrieved or derived from the English Housing Survey – Fuel Poverty datasets (EHS-FP) \cite{UKDA9243}, covering the period from 2015 to 2018.}\\

\endfirsthead

\multicolumn{2}{l}{\textit{Table \thetable\ (continued)}} \\
\hline
\textbf{Variable name} & \textbf{Var. symbol} & \textbf{Unit} & \textbf{Values} & \textbf{Description} \\
\hline
\endhead
\hline
\textbf{Variable name} & \textbf{Var. symbol} & \textbf{Unit} & \textbf{Values} & \textbf{Notes} \\
\hline
Walls insulation & $X$ & -- & \makecell[l]{True, \\ False} & Values obtained by collapsing the four-value variable ``WallType'' in the EHS-FP dataset \cite{UKDA9243}, which indicates whether external walls are cavity or solid and whether it is insulated or uninsulated. \\
\hline
\makecell[l]{Gas consumption \\ for space heating} & $Y_0$ & kWh/year & \makecell[l]{Real-valued} & Values are obtained from space heating gas cost, $v_1$, and gas prices, $v_9$, i.e. $y_0 = v_1 / v_9$. \\
\hline
\makecell[l]{Energy burden} & $W$ & £/£ & \makecell[l]{Real-valued \\ $\in [0,1]$} & This variable represents the percentage of annual household income spent on gas and electricity costs. Values are obtained from space heating cost plus other gas and electricity costs, i.e. $w=(v_1+v_{10})/v_7$.\\
\hline
Dwelling type & $V_0$ & -- & \makecell[l]{End terrace, \\ Mid terrace, \\ Semi detached, \\ Detached, \\ Purpose built flat, \\ Converted flat}  & \\
\hline
\makecell[l]{Space heating \\gas cost} & $V_1$ & £/year & \makecell[l]{Real-valued}  & \\
\hline
\makecell[l]{Tenancy \\} & $V_2$ & -- & \makecell[l]{Owner-occupied, \\ Private rented, \\ Local authority, \\ Housing association}  & \\
\hline
\makecell[l]{Dwelling age} & $V_3$ & -- & \makecell[l]{$< 1850$, \\ 1850-1899, \\ 1900-1918, \\ 1919-1944, \\ 1945-1964, \\ 1965-1974, \\ 1975-1980, \\ 1981-1990, \\ $> 1990$}  & Age of the oldest part of the building, indicating the period in which it was built. \\
\hline
\makecell[l]{Under-occupancy \\} & $V_4$ & -- & \makecell[l]{True, \\ False}  & \\
\hline
\makecell[l]{Household size \\} & $V_5$ & -- & \makecell[l]{1, 2, 3, 4, \\ 5 or more}  & Number of members of the household. \\
\hline
\makecell[l]{Floor area \\} & $V_6$ & m\textsuperscript{2} & \makecell[l]{$< 50$, \\ 50-69, \\ 70-89, \\ 90-109, \\ $> 109$ }  & \\
\hline
\makecell[l]{\\ Income \\} & $V_7$ & £/year & \makecell[l]{Real-valued}  & Annual full household income. This is calculated using net income, which includes housing benefit, Support for Mortgage Interest (SMI), Mortgage Payment Protection Insurance (MPPI), and net council tax payments. It accounts for the total household income from all sources, including benefits, savings, and investments \cite{UKDA9243}. \\
\hline
\makecell[l]{Household \\ composition } & $V_8$ & -- & \makecell[l]{- Couple under 60 without \\ dependent child(ren), \\- Couple aged 60 or over \\ without dependent \\ child(ren), \\- Couple with dependent \\ child(ren), \\- Lone parent with \\ dependent child(ren), \\- Other multi-person \\ household, \\- One person under 60, \\- One person aged 60  \\ or over, \\- N/A}  & \\
\hline
\makecell[l]{\\ Gas price \\} & $V_9$ & £/kWh & \makecell[l]{Real-valued}  & Retail gas prices per unit of energy (for different years) are derived from domestic gas bills, categorised by method of payment: Direct Debit, Standard Credit, and Prepayment. These domestic gas bill data are published by the UK Govt. -- Department for Energy Security and Net Zero \cite{Gas_billsData}. Gas prices are then matched to household units in the EHS-FP datasets based on the value of the ``gasmop'' variable. \\
\hline
\makecell[l]{Other gas and \\ electricity costs} & $V_{10}$ & £/year & \makecell[l]{Real-valued}  & The variable includes costs for lights and appliances use and cost for cooking, i.e. respectively ``litecost'' and ``cookcost'' in the EHS-FP datasets \cite{UKDA9243}. \\
\hline
\end{longtable}
}
\end{center}

\newpage

\subsection{Direct causal relations}\label{subsec_direct_causal_relations}

\begin{center}
\label{Axiomatic_causes}
{\footnotesize
\begin{longtable}{|p{4.8cm}|p{1.2cm}|p{7.5cm}|}

\caption{\small Axiomatic causal relationships between variables in the graph $\mathcal{G}$ (shown in Figures \ref{fig:main_graphs}-a and \ref{fig:subgraphs} (top-left)). Causal relations are elucidated in terms of the direct causal mechanisms between variable pairs, thus describing the assumed mechanism of how, \emph{ceteris paribus}, each parent variable causally affects its corresponding child(ren) variable(s). }\\

\endfirsthead

\multicolumn{2}{l}{\textit{Table \thetable\ (continued)}} \\
\hline
\textbf{Parent $\rightarrow$ child } & \textbf{Var(s). symbols} & \textbf{Rationale of causal mechanism} \\
\hline
\endhead

\hline
\textbf{Parent $\rightarrow$ child } & \textbf{Var(s). symbols} & \textbf{Rationale of causal mechanism} \\
\hline
Walls insulation $\rightarrow$ Gas consumption for space heating & $X \rightarrow Y_0$ & The presence of insulation on external walls reduces heat loss, improving the dwelling’s thermal efficiency and lowering the energy needed for heating, leading to a reduction in gas consumption for space heating. \\
\hline
Other efficiency interventions $\rightarrow$ Gas consumption for space heating & $U_3 \rightarrow Y_0$ & Energy efficiency interventions other than wall insulation are assumed to reduce gas consumption for space heating via physical mechanisms, whether through increased efficiency of the external fabric or because of the replacement or upgrade of newer, more energy efficient, heating systems or devices. \\
\hline
Other efficiency interventions $\rightarrow$ Space heating gas cost & $U_3 \rightarrow V_1$ & The same causal mechanism described for $U_3 \rightarrow Y_0$ also applies to Space heating gas cost. \\
\hline
Walls insulation $\rightarrow$ Space heating gas cost & $X \rightarrow V_1$ & External walls insulation improves the dwelling’s thermal efficiency, which decreases household’s cost expenditures required for space heating. \\
\hline
Energy burden $\rightarrow$ Gas consumption for space heating & $W \rightarrow Y_0$ & As heating cost takes up a larger portion of household income (whether due to a decrease in income or to rising gas prices) households may adjust their living habits to reduce gas consumption. This can include lowering thermostat settings, heating only specific rooms, or adopting energy-saving practices such as wearing extra layers or using space heaters selectively. \\
\hline
Space heating gas cost $\rightarrow$ Energy burden & $V_1 \rightarrow W$ & Higher (lower) household expenditure for space heating increases (decreases) the proportion of income spent on gas, thus contributing to raise (lower) the overall energy burden.  \\
\hline
Other gas and electricity cost $\rightarrow$ Energy burden & $V_{10} \rightarrow W$ & Higher (lower) gas and electricity cost (other than for space heating) have an increasing (decreasing) effect on the proportion of income spent for energy, thus contributing to raise (lower) the overall energy burden.  \\
\hline
Income $\rightarrow$ Energy burden & $V_7 \rightarrow W$ & Higher (lower) household income reduces (increases) heating energy burden by decreasing (increasing) the relative percentage share of income spent on heating. \\
\hline
Education attainment $\rightarrow$ Income & $U_2 \rightarrow V_7$ & Higher education levels generally lead to greater job opportunities, higher-paying positions, and increased earning potential, resulting in higher (earned) incomes for the household. \\
\hline
Education attainment $\rightarrow$ Gas consumption for space heating & $U_2 \rightarrow Y_0$ & The causal relationship between household educational attainment and heating energy use is mediated by energy literacy and climate change awareness (two factors not explicitly modelled here). More educated households tend to have greater knowledge of energy efficiency and conservation strategies, enabling them to adopt behaviours that reduce heating energy use. Environmental concern is another key mediating factor. Higher education is often linked to a greater awareness of climate change and environmental issues, which can further motivate households to adopt energy-saving behaviours. Educated individuals are more likely to recognise the link between personal energy use and carbon emissions, leading them to consciously reduce their heating habits to minimise their environmental footprint.  \\
\hline
Education attainment $\rightarrow$ Space heating gas cost & $U_2 \rightarrow V_1$  & The same causal mechanism that explains how household education attainment affects heating energy (gas) consumption also explains its impact on space heating costs. Note: gas consumption and cost are strongly correlated, with the slope of the linear relationship representing the retail price of gas per unit of energy [£/kWh]. \\
\hline
Other behavioural factors $\rightarrow$ Gas consumption for space heating & $U_5 \rightarrow Y_0$ & As for education attainment, we acknowledge the existence of other variables potentially driving energy consumption that are essentially behavioural in nature, such as awareness of energy efficiency, environmental values, or habitual energy-saving practices. These factors are not directly observable but are likely to play a role in shaping how individuals use energy within their homes. To account for their potential influence on gas consumption, we incorporate them into the graphical model by collapsing them into a single unobserved latent variable. \\
\hline
Other behavioural factors $\rightarrow$ Space heating gas cost & $U_5 \rightarrow V_1$ & The same reasoning that supports the influence of behavioural factors on gas consumption ($U_5 \rightarrow Y_0$) also also applies to gas cost expenditures. Since gas cost is directly determined by the quantity of gas consumed and the retail price of gas, and given that the retail price is exogenous to household behaviour, any behavioural influence on gas cost must operate indirectly via its effect on consumption. Therefore, the latent behavioural variable $U_5$ also affects $V_1$ through the same underlying mechanisms.    \\
\hline
Gas price $\rightarrow$ Household gas cost & $V_9 \rightarrow V_1$ & Any change in the retail price of gas per unit of energy generated directly affects gas cost expenditures faced by the household, assuming the same level of gas consumption.  \\
\hline
Dwelling age $\rightarrow$ Walls insulation & $V_3 \rightarrow X$ & The causal mechanism between dwelling age and external wall insulation operates through differences in construction standards and retrofit likelihood. Older dwellings were built with less stringent thermal regulations and may lack insulation unless they have undergone a retrofit, thus introducing higher variability in insulation presence for these group. In contrast, newer dwellings are more likely to include external walls insulation by default due to progressively stricter building regulations mandating higher energy efficiency standards. \\
\hline
Dwelling type $\rightarrow$ Gas consumption. & $V_0 \rightarrow Y_0$ & The influencing mechanism through which dwelling type affects gas consumption for space heating is geometrical in nature. Specifically, it is related to the amount of externally exposed surface area per unit of internal volume (often measured in term of \emph{compactness} ratio = surface area / volume). As such, detached homes typically require more energy for heating than semi-detached or terraced houses. Apartment flats tend to have the lowest amount of externally exposed surface area per unit of internal volume. \\
\hline
Dwelling type $\rightarrow$ Space heating gas cost & $V_0 \rightarrow V_1$ & The same causal mechanism that explains how dwelling type affects gas consumption ($V_0 \rightarrow Y_0$) applies to gas cost expenditures faced by the household.  \\
\hline
Income $\rightarrow$ Tenancy & $V_7 \rightarrow V_2$ & Higher household income increases the likelihood of home-ownership, as higher earners can afford down payments, qualify for mortgages, and benefit from long-term equity accumulation. In contrast, lower-income households are more likely to rent on the private market or from local authorities (or housing associations) due to affordability constraints and limited access to credit. \\
\hline
Tenancy $\rightarrow$ Walls insulation & $ V_2 \rightarrow X$ & Owner-occupied dwellings are more likely to feature insulated walls due to homeowners’ stronger financial motivation to minimise long-term energy expenses and their full autonomy over property improvements. In contrast, privately rented properties are less likely to have insulated walls due to the split incentive issue, where landlords have little direct benefit from energy efficiency upgrades since tenants, rather than landlords, typically bear the energy costs. As a result, private landlords may avoid investing in external wall insulation, especially in lower-cost rental markets where affordability is a key concern for tenants.

However, renters in social housing ---including those in local authority-managed housing and housing association properties--- are more likely to live in homes with insulated external walls. This is largely due to the historical uptake of insulation in the social housing sector, driven by government-led energy efficiency programs, stricter building regulations, and policy initiatives aimed at improving housing quality for lower-income residents. Social landlords, unlike private landlords, often have long-term investment strategies and access to public funding for large-scale retrofit programs, making insulation upgrades more common in these properties. \\
\hline
Under-occupancy $\rightarrow$ Gas consumption for space heating & $V_4 \rightarrow Y_0$ & Under-occupancy leads to higher gas use per person because fewer people living in a house than it was designed for means that more volume of air is heated per person, assuming all else is equal. However, its impact on the total (absolute) energy use per dwelling may to result in a reduction of energy use, since occupants may adjust their heating behaviour by turning off radiators in unused rooms or using zoned heating systems to limit energy use to just the areas they occupy.  \\
\hline
Under-occupancy $\rightarrow$ Space heating gas cost & $V_4 \rightarrow V_1$ & The same causal mechanism described above for $V_4 \rightarrow Y_0$ also applies to Space heating gas cost. \\
\hline
Household size $\rightarrow$ Under-occupancy & $ V_5 \rightarrow V_4$ & Household size directly influences whether a dwelling is under-occupied because as the number of occupants decreases (e.g. due to life events like children moving out, separation, or else) the amount of space per person increases.  \\
\hline
Floor area $ \rightarrow $ Under occupancy & $V_6 \rightarrow V_4$ & The floor area of a dwelling influences whether it is considered under-occupied by definition because under-occupancy is measured based on the floor area (or number of rooms) relative to the number of occupants.   \\
\hline
Floor area $\rightarrow$ Gas consumption for space heating & $V_6 \rightarrow Y_0$ & Dwelling floor area causally affects energy (gas) use for space heating because larger homes have a greater volume of air which directly increases the amount of energy required to achieve a target comfort temperature. In relative terms however, larger homes tend to require less heating energy per unit of floor area due to an increase of \emph{compactness}. As floor area increases, the ratio of external surface area (walls, windows, roof) to the internal volume decreases, reducing heat loss per unit of floor space. \\
\hline
Floor area $\rightarrow$ Space heating gas cost & $V_6 \rightarrow V_1$ & The same causal mechanism described above for $V_6 \rightarrow Y_0$ also applies to space heating gas cost. \\
\hline
Location $\rightarrow$ Floor area & $ U_1 \rightarrow V_6 $ & The dwelling's geographic location influences floor area because rural areas typically offer more space allowing for larger dwellings at lower costs, while urban areas, with higher demand and limited space, tend to have smaller dwellings.  \\
\hline
Location $\rightarrow$ Income & $ U_1 \rightarrow V_7 $ & Geographic location affects household income because urban areas typically offer more job opportunities, higher wages, and access to diverse industries, while rural areas may have fewer employment opportunities and hence a lower income, on average. \\
\hline
Income $\rightarrow$ Floor area & $V_7 \rightarrow V_6$ & A higher-income household it is more likely to live in a dwelling with lager floor area, as wealthier households can afford bigger homes, whether through purchasing, renting, or extending existing properties. 
Note how the causal association between higher-income households and increased heating energy consumption/expenditure it is captured here via the two mediating variables \emph{dwelling floor area} and \emph{under-occupancy}, that is, wealthier households tend to have higher heating costs because (among other reasons) they occupy larger, and more often under-utilised homes.
 \\
\hline
Household composition $\rightarrow$ Income & $V_8 \rightarrow V_7$ & Household composition strongly influences income due to differences in earning potential and dependency ratios. Couples without children, particularly those under 60, often have higher combined incomes since both partners are likely to be employed full-time. Couples with children may also earn well, but their disposable income can be reduced by childcare costs and one partner potentially working part-time. In contrast, single-person households—especially lone parents or young professionals—tend to have lower total income, often due to a single earner and possible part-time employment. Older households, whether individuals or couples, generally experience reduced income as they transition into retirement. Multi-person households may benefit from multiple income sources, leading to a higher overall household income.  \\
\hline
Household composition $\rightarrow$ Gas consumption for space heating & $V_8 \rightarrow Y_0$ & Household composition has a causal impact on heating energy consumption due to differences in occupancy patterns and needs. Household with children or occupants over 60 may require warmer indoor temperature due to lower tolerance, thus driving up heating energy use. Households where members are more likely to stay at home for longer periods, e.g. couples aged 60 or over, or lone parents with young children, tend to use more heating energy overall, as their homes need to remain warm throughout the day. Conversely, working-age couples (especially with no dependent children) tend to have lower heating energy demand, as heating is only needed during morning and evening hours during the week.  \\
\hline
Household composition $\rightarrow$ Space heating gas cost & $V_8 \rightarrow V_1$ & The same reasoning that supports the causal effect of household composition on gas consumption ($V_8 \rightarrow Y_0$) applies to gas expenditures for the household.  \\
\hline
Outdoor temperature $\rightarrow$ Gas consumption for space heating & $U_4 \rightarrow Y_0$ & Outdoor temperature affects gas consumption for space heating, as colder temperatures increase heat loss from the building, thus requiring more energy input to achieve and maintain adequate indoor temperatures.  \\
\hline
Outdoor temperature $\rightarrow$ Space heating gas cost & $U_4 \rightarrow V_1$ & Outdoor temperature influences household heating cost, as lower external temperatures increase heat loss from the dwelling, thus requiring the household to purchase more energy in order to maintain indoor temperature at adequate levels.  \\
\hline

\end{longtable}
}
\end{center}

\newpage

\subsection{p-value computation for placebo test}\label{SM_p_val_placebo}
As described in section \ref{section_placebo_tretment_test}, we assess the statistical significance of the estimated treatment effect $ATE_{\mathcal{G}}$ by comparing it to a statistical distribution of placebo treatment effects $ATE_{i,\text{placebo}}$, generated from $n = 1600$ datasets iterations of randomised treatment permutations.

The p-value is then computed as:

\begin{equation} 
\label{p-value_eq_a}
\text{p-value} = \frac{1+\sum_{i=1}^{n} 1(|ATE_{i, \text{placebo}} | \geq |ATE_{\mathcal{G}}|)}{1+n} 
\end{equation}

Here, $1(\cdot)$ is the indicator function counting how many $i$ instances of $ATE_{\text{placebo}}$ are as extreme (or more extreme) than the estimated true effect $ATE_{\mathcal{G}}$. As such, the computed p-value represents the probability of observing a treatment effect as extreme as $ATE_{\mathcal{G}}$ under the null hypothesis that the treatment is causally unrelated to the outcome. A low p-value therefore provides strong evidence that the observed effect is unlikely to have occurred under random (placebo) assignment.

Note how the numerator and denominator in Eq. (\ref{p-value_eq_a}) are offset by 1 to avoid zero p-values in finite samples, thus ensuring a conservative p-value estimate.

\subsection{p-value computation for subsample test}\label{SM_p_val_subsample}

To compute a two-sided p-value for the subsample refutation test, we assess how extreme is the estimated true effect ($ATE_{\mathcal{G}}$) using the full training dataset compared to the treatment effects $ATE_{i, \text{sub}}$ estimated $n$ times from random subsamples of the training dataset:

\begin{equation} 
\label{p-value_eq_b}
\text{p-value} = \frac{1+\sum_{i=1}^{n} 1(|ATE_{i, \text{sub}} - ATE^*_{i, \text{sub}} | \geq |ATE_{\mathcal{G}} - ATE^*_{i, \text{sub}}|)}{1+n} 
\end{equation}
where: 
\begin{itemize}
  \item[-] $ATE_{i, \text{sub}}=$ average treatment effect computed using the $i$-th random subsample dataset.
  \item[-] $ATE^*_{i, \text{sub}}=$ mean value of average treatment effects from all subsamples.
  \item[-] $ATE_{\mathcal{G}}=$ average treatment effect estimated from the full sample dataset.
  \item[-] $1(\cdot)=$ indicator function counting how many instances $i$ of $ATE_{ \text{sub}}$ are as extreme as $ATE_{\mathcal{G}}$ from the mean value of average treatment effects from all subsamples. 
\end{itemize}

As such, the test effectively quantifies how well the full-sample effect $ATE_{\mathcal{G}}$ is within the subsample distribution. If it lies somewhere near the mean value, a large p-value will result from Eq. (\ref{p-value_eq_b}), whereas a low p-value will result from the same if $ATE_{\mathcal{G}}$ it's unusually far from $ATE^*_{i, \text{sub}}$, thus indicating instability of the effect estimate.

\subsection{Illustrative example of ATE and CATE calculation}\label{ATECATE_dummy_exemple}

To bridge the gap between theoretical formulae (Eqs. (\ref{eq:expectation}-\ref{EQ:ATE_CATE_G}) in the main text) and practical implementation, we provide a simple worked example of ATE and CATE using dummy data. Assuming the outcome variable $Y_0$ (gas consumption) is discretised into three levels $y_0 \in \{5,10,15 \}$ kWh/year and with post-intervention probability distributions as shown in Table \ref{tab:probabilities_y0}.

\begin{table}[h!]
\caption{Probabilities of $y_0$ under interventions on $X$}
\label{tab:probabilities_y0}
\centering
\begin{tabular}{c c c}
\hline
$y_0$ [kWh/year] & $P(y_0 \mid do(X=\text{false}))$ & $P(y \mid do(X=\text{true}))$ \\
\hline
5  & 0.20 & 0.40 \\
10 & 0.50 & 0.40 \\
15 & 0.30 & 0.20 \\
\hline
\end{tabular}
\end{table}

Using the first of Eqs. (\ref{eq:expectation}):

\begin{equation}
\begin{split}
    & \mathbb{E}(Y_0 \mid do(X=\text{false})) =5(0.20)+10(0.50)+15(0.30)=10.5  \\
    & \mathbb{E}(Y_0 \mid do(X=\text{true})) =5(0.40)+10(0.40)+15(0.20)=9 \\
\end{split}
\end{equation}

Thus, the average treatment effect is (as per the first of Eqs. (\ref{EQ:ATE_CATE_G})): 

\begin{equation}
ATE = 9.0 - 10.5 = -1.5 \ \text{kWh/year}
\end{equation}
indicating that wall insulation reduces average gas consumption by 1.5 kWh/year. We can also examine heterogeneity by conditioning on energy burden $W$. Assuming, for illustration, $W$ is a binary (low; high) variable, and the post-intervention probability distributions conditional on $W=\text{high}$ are as shown in Table \ref{tab:probabilities_y0_given_w}, using the second of Eqs. (\ref{eq:expectation}) we have:

\begin{equation}
\begin{split}
    & \mathbb{E}(Y_0 \mid do(X=\text{false}), W=\text{high}) =5(0.10)+10(0.60)+15(0.30)=11.0  \\
    & \mathbb{E}(Y_0 \mid do(X=\text{true}), W=\text{high}) =5(0.20)+10(0.50)+15(0.30)=10.5 \\
\end{split}
\end{equation}

Thus, the average treatment effect conditional on $W=\text{high}$ is (as per the second of Eqs. (\ref{EQ:ATE_CATE_G})): 

\begin{equation}
CATE_{W=\text{high}} = 10.5 - 11.0 = -0.5 \ \text{kWh/year}
\end{equation}
indicating that wall insulation reduces average gas consumption by 0.5 kWh/year for this specific subgroup.

\begin{table}[h!]
\caption{Probabilities of $y_0$ under interventions on $X$ conditional on $W$.}
\label{tab:probabilities_y0_given_w}
\centering
\begin{tabular}{c c c}
\hline
$y_0$ [kWh/year] & $P(y_0 \mid do(X=\text{false}), W= \text{high})$ & $P(y \mid do(X=\text{true}), W= \text{high})$ \\
\hline
5  & 0.10 & 0.20 \\
10 & 0.60 & 0.50 \\
15 & 0.30 & 0.30 \\
\hline
\end{tabular}
\end{table}

\end{document}